\documentclass[nohyperref]{article}

\usepackage{microtype}
\usepackage{graphicx}
\usepackage{caption}
\usepackage{subcaption}
\usepackage{booktabs} % for professional tables

\usepackage{hyperref}

\usepackage[accepted]{icml2022}

\usepackage{amsmath}
\usepackage{amssymb}
\usepackage{mathtools}
\usepackage{amsthm}
\usepackage[T1]{fontenc}
\usepackage{multirow}
\usepackage{makecell}
\usepackage[capitalize,noabbrev]{cleveref}

\theoremstyle{plain}

\theoremstyle{definition}

\theoremstyle{remark}

\usepackage[textsize=tiny]{todonotes}

\icmltitlerunning{Submission and Formatting Instructions for ICML 2022}

\begin{document}

\twocolumn[
\icmltitle{Collaboration of Experts: Achieving 80\% \\ Top-1 Accuracy on ImageNet with 100M FLOPs}



\icmlsetsymbol{equal}{*}

\begin{icmlauthorlist}
\icmlauthor{Yikang Zhang}{comp}
\icmlauthor{Zhuo Chen}{comp}
\icmlauthor{Zhao Zhong}{comp}
\end{icmlauthorlist}

\icmlaffiliation{comp}{Huawei, Beijing, China}

\icmlcorrespondingauthor{Yikang Zhang}{zhangyikang7@huawei.com}
\vskip 0.3in
]

\printAffiliationsAndNotice{}

\begin{abstract}
In this paper, we propose a Collaboration of Experts (CoE) framework to assemble the expertise of multiple networks towards a common goal. Each expert is an individual network with expertise on a unique portion of the dataset, contributing to the collective capacity. Given a  sample, delegator selects an expert and simultaneously outputs a rough prediction to trigger potential early termination. For each model in CoE, we propose a novel training algorithm with two major components: weight generation module (WGM) and label generation module (LGM). It fulfills the co-adaptation of experts and delegator. WGM partitions the training data into portions based on delegator via solving a balanced transportation problem, then impels each expert to focus on one portion by reweighting the losses. LGM generates the label to constitute the loss of delegator for expert selection. CoE achieves the state-of-the-art performance on ImageNet, 80.7\% top-1 accuracy with 194M FLOPs. Combined with PWLU and CondConv, CoE further boosts the accuracy to 80.0\% with only 100M FLOPs for the first time. Furthermore, experiment results on the translation task also demonstrate the strong generalizability of CoE. CoE is hardware-friendly, yielding a 3$\sim$6x acceleration compared with existing conditional computation approaches.
\end{abstract}

\section{Introduction}

From simple systems to complicated ones, the accomplishment of various tasks relies on the collaboration of multiple individuals. Similarly, a wise combination of models with different properties could yield improved performance on a specific task compared to only deploying one individual model. There are many approaches for model collaboration, among which ensemble learning~\cite{hansen1990neural, wen2020batchensemble, wenzel2020hyperparameter} is a popular one. Ensemble learning uses a consensus scheme to decide the collective result by vote. However, it requires multiple forward passes, leading to a significant runtime cost. MIMO~\cite{havasi2020training} draws inspiration from model sparsity~\cite{frankle2018lottery} and tries to ensemble several subnetworks within one regular network. It only needs one single forward pass of the regular network but is incompatible with compact models. Conditional computation methods~\cite{cheng2020instanas, yan2015hd} adopt the delegation scheme for model collaboration, conditionally assigning one or several, rather than all models, to make the prediction. Some recently proposed conditional computation methods~\cite{yang2019condconv,zhang2021basisnet} have achieved remarkable performance based on dynamic convolution. Nonetheless, they usually have high memory access cost (MAC) and a low degree of parallelism, increasing the real latency~\cite{ma2018shufflenet}.

Motivated by this, we propose the Collaboration of Experts (CoE) framework to both eliminate the need for multiple forward passes and keep hardware-friendly. CoE consists of one delegator and multiple experts. Firstly, delegator gives a rough prediction and makes the expert selection.  If the rough prediction is unreliable, the selected expert will make the refined prediction. Otherwise, the procedure will be early terminated to save FLOPs. Moreover, we only need to load the selected expert into memory, thus keeping the ratio of MAC to FLOPs as a constant. By contrast,  dynamic convolution methods~\cite{zhang2020dynet, zhang2021basisnet} need to load a large number of parameters, namely basis models or experts, to synthesize the input-dependent ones. It enlarges MAC and reduces the degree of parallelism, resulting in a significant deceleration.

\begin{figure}
	\centering
	\includegraphics[width=\linewidth]{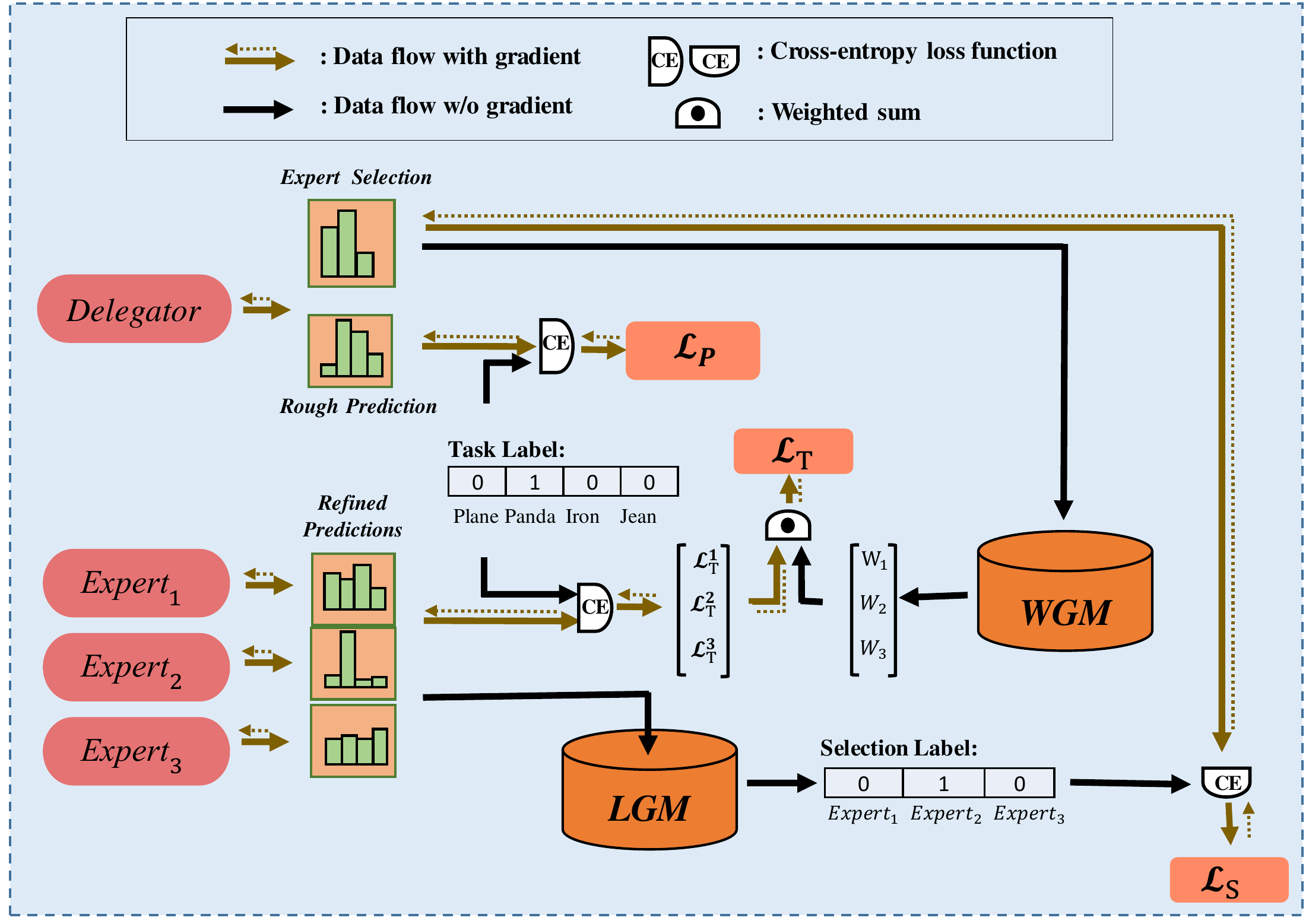}
	\caption{Training framework of CoE with three experts. $\mathcal{L}_T^1$, $\mathcal{L}_T^2$ and $\mathcal{L}_T^3$ are the expert losses, which are reweighted by WGM with weights $W_1$, $W_2$ and $W_3$. LGM generates the label to constitute the loss $\mathcal{L}_S$ for expert selection. $\mathcal{L}_P$ is the cross-entropy loss for the rough prediction of delegator.
	}
	\label{framework}
\end{figure}

\begin{figure}
	\centering
	\includegraphics[width=0.7\linewidth]{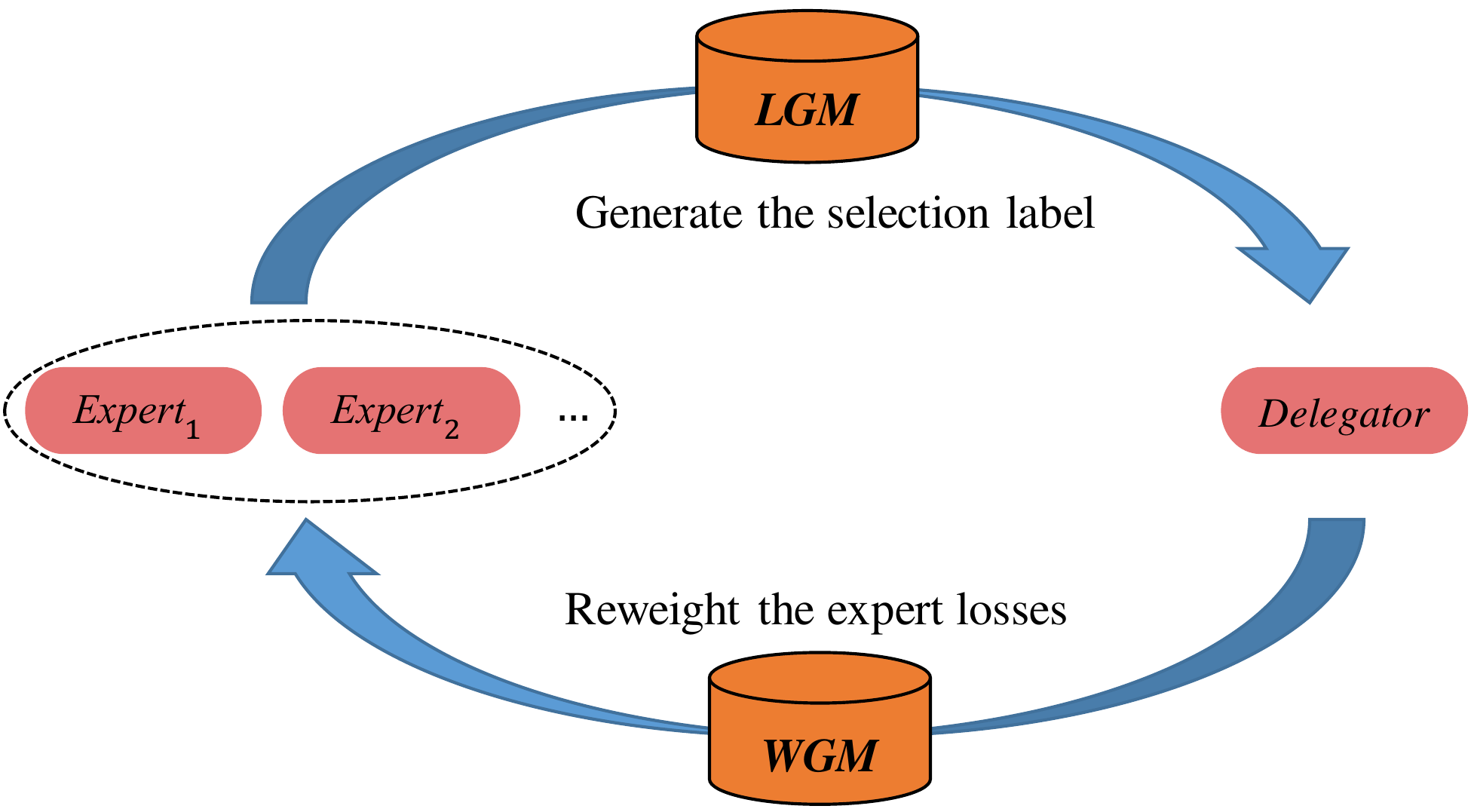}
	\caption{The co-adaptation of experts and delegator. LGM generates the selection label which indicates the suitable expert to supervise delegator for expert selection. WGM partitions the training data into portions based on delegator and impels each expert to focus on one portion by reweighting the losses.}
	\label{coadaptation}
\end{figure}

To make each model in CoE play its role, we propose a novel training algorithm (as shown in Fig.\ref{framework}) which consists of two major components: weight generation module (WGM) and label generation module (LGM). LGM generates the label (selection label) to constitute the loss of delegator for expert selection. Selection label is a one-hot vector, indicating the suitable expert for each given input. It is obtained by solving a balanced transportation problem (BTP, \citealt{shore1970transportation}). Based on delegator, WGM partition the training data into portions by maximizing the summed selection probability via solving BTP as well. Then expert losses are reweighted so that each expert can focus on one portion. As shown in Fig.\ref{coadaptation}, this fulfills the co-adaptation of experts and delegator. The co-adaptation manner makes CoE generalize well to the validation set. Due to the random initialization of experts, selection labels are irregular in the early training stage. Nonetheless, delegator tends to learn generalizable patterns first, since networks learn gradually more complex hypotheses during training~\cite{pmlr-v70-arpit17a}. Therefore, WGM can partition the training data into portions based on generalizable patterns with delegator as the bridge. It makes selection labels more regular in return, thus delegator avoids overfitting to the irregular labels. 

We conduct the main experiments on ImageNet classification task. CoE achieves 78.2/80.7\% top-1 accuracy with only 100/194M FLOPs, while the accuracy for ensembled models~\citep{hansen1990neural} is only 79.6\% with 920M FLOPs. Compared with the widely-used gate-value-based optimization method~\citep{shazeer2017outrageously, fedus2021switch}, our proposed training algorithm improves 1.2\% accuracy for CoE, indicating the training effectiveness. Compared with dynamic network approaches, CoE is more hardware-friendly. It not only outperforms the SOTA dynamic method BasisNet which achieves 80.0\% accuracy with 198M FLOPs~\citep{zhang2021basisnet}, but also accomplishes a 3.1x speedup on hardware. Besides, CoE can be equipped with CondConv and further improve the accuracy to 79.2/81.5\% with 102/214M FLOPs. Moreover, we further boost the accuracy to 80.0\% with only 100M FLOPs for the first time by using PWLU activation function~\citep{DBLP:journals/corr/abs-2104-03693}. Experiment results on the translation task also demonstrate the strong generalizability of CoE.

The contributions of this paper can be summarized as follows:
\begin{itemize}
	\item We propose a collaboration framework named Collaboration of Experts (CoE). The core advantage of it is the inference efficiency. Compared with other conditional computation methods, CoE has low memory access cost and a high degree of parallelism, which are two important factors for real latency.
	\item We present a novel optimization strategy for CoE that fulfills the co-adaptation of experts and delegator. Experiment results demonstrate its superiority over the widely-used gate-value-based optimization method.
	\item CoE updates the state-of-the-art on ImageNet for mobile setting, achieving 80.7\% top-1 accuracy on ImageNet with less than 200M FLOPs for the first time as far as we know.
	\item Since the expert selection is done at the model level, CoE can take advantage of existing techniques like conditional convolution and PWLU activation function to push the performance to a new level, namely, 80.0\% accuracy on ImageNet with only 100M FLOPs.
\end{itemize}

\section{Related Work}
\subsection{Ensemble Learning and {Model Selection}}

Ensemble learning ~\citep{hansen1990neural} aims at combining the predictions from several models to get a more robust one. Some recently proposed literatures~\citep{wen2020batchensemble,wenzel2020hyperparameter} demonstrate that significant gains can be achieved with negligible additional parameters compared to the original model. However, these methods still require multiple (typically, 4-10) forward passes for prediction, leading to a significant runtime cost. Differently, CoE utilizes a delegator to select only one expert for the refined prediction, thus at most two forward passes are needed. MIMO~\citep{havasi2020training} draws inspiration from model sparsity~\citep{frankle2018lottery} and holds the view that multiple independent subnetworks can be concurrently trained within one regular network because of the heavy parameter redundancy. Therefore, those subnetworks can be ensembled with a single forward pass of the regular model. However, MIMO cannot be applied to compact models which have already been pruned or the ones constructed by AutoML methods ~\citep{cai2019once,8578355}. It is because these models have few redundant parameters. By contrast, CoE is free from the compactness of experts since expert selection is done at the model level. Recently, some works about model selection are proposed~\citep{Li_ranking_2021_CVPR, pmlr-v139-you21b}. These methods are concerned with ranking a number of pre-trained models and finding the one transfers best to a downstream task of interest. Therefore, they select models task-wisely. By contrast, CoE aims at improving the task performance via selecting the most suitable expert for each sample instance-wisely. 

\subsection{Dynamic Networks}

Dynamic networks achieve high performance with low computation cost by conditionally varying the network parameters~\citep{zhang2020dynet,yang2019condconv} or network architectures~\citep{yuan2020dynamic}. 
HD-CNN~\citep{yan2015hd} and HydraNet~\citep{mullapudi2018hydranets} select branches based on the category, they cluster all categories into n groups, where n is the number of branches. While CoE learns the model selection pattern automatically, it can be based on any property, rather than limited to the category. MoE~\citep{shazeer2017outrageously} and Switch Transformer~\citep{fedus2021switch} select experts at the layer level with a router. The output feature of each expert will be scaled with the gate-value predicted by router, thus router becomes trainable. This gate-value-based optimization manner is heuristic while CoE trains the delegator more reasonably. Since expert selection for CoE is done across models, we can use protocols like True Class Probability (TCP, \citealt{corbiere2019addressing}) to measure the suitability of each expert without bias. Based on expert suitabilities, the labels to supervise delegator for expert selection can be obtained. Additionally, CoE takes more advantage of conditional computation as the whole network is selected, rather than only some particular layers. The recently proposed Dynamic Convolution methods~\citep{zhang2020dynet,yang2019condconv,chen2020dynamic} share the similar idea and achieve remarkable performance with low FLOPs but high latency. It is because these methods need to load many basis models or experts to synthesize the dynamic parameters, causing high MAC and low degree of parallelism~\citep{ma2018shufflenet}. By contrast, CoE only needs to load the selected expert into memory, avoiding these problems. Finally yet importantly, batch processing is an important method to enhance the degree of parallelism. Because of the input-dependent parameters~\citep{shazeer2017outrageously, zhang2021basisnet} or architectures~\citep{yuan2020dynamic}, these methods cannot process samples in batch. Differently, CoE supports batch processing because the number of experts is limited and each one of them corresponds to many test samples.

\section{Method}

\begin{figure}[tbh]
	\centering
	\includegraphics[width=0.6\linewidth]{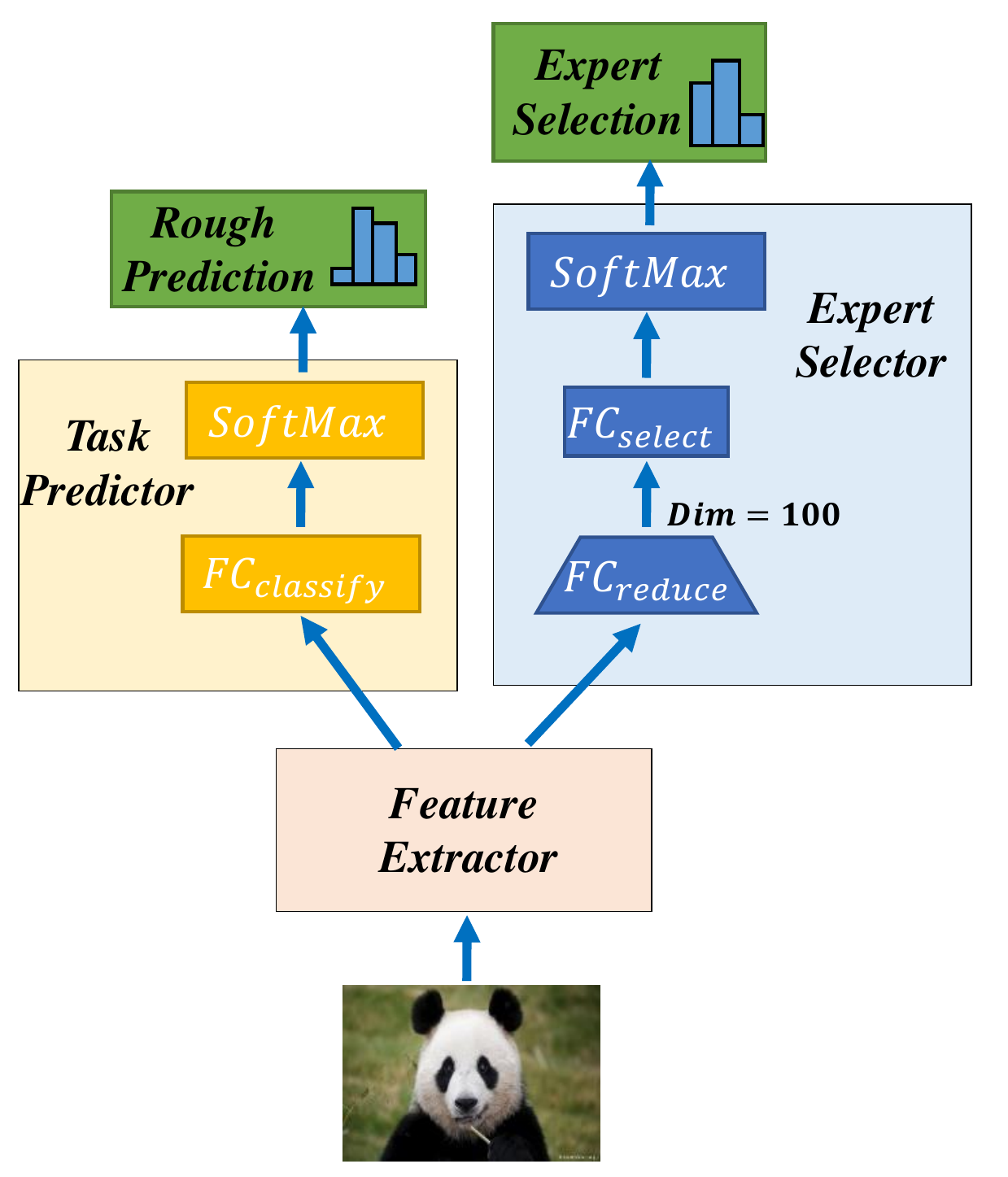}
	\caption{The architecture of delegator. Task predictor consists
		of one fully connected layer and a SoftMax layer. Expert
		selector consists of two fully connected layers with the
		hidden dim as 100, followed by a SoftMax layer.}
	\label{generalist}
\end{figure}

CoE consists of a delegator and $n$ experts, a total of $n+1$ individual neural networks. Given a sample, delegator will select an expert and simultaneously output a rough prediction to trigger potential early termination.  Since the inference of delegator is conducted all the time, we prefer to make delegator more lightweight than expert. Delegator consists of three modules: feature extractor, task predictor and expert selector as shown in Fig.\ref{generalist}.  Based on the feature derived from feature extractor, task predictor and expert selector output the probabilities for classification and expert selection respectively. In the following subsections, we will describe the inference procedure and training strategy of CoE comprehensively. The number of samples and experts are denoted as $m$ and $n$.

\subsection{{Inference Procedure}}\label{inference_procedure}

\begin{figure}[tbh]
	\centering
	\includegraphics[width=\linewidth]{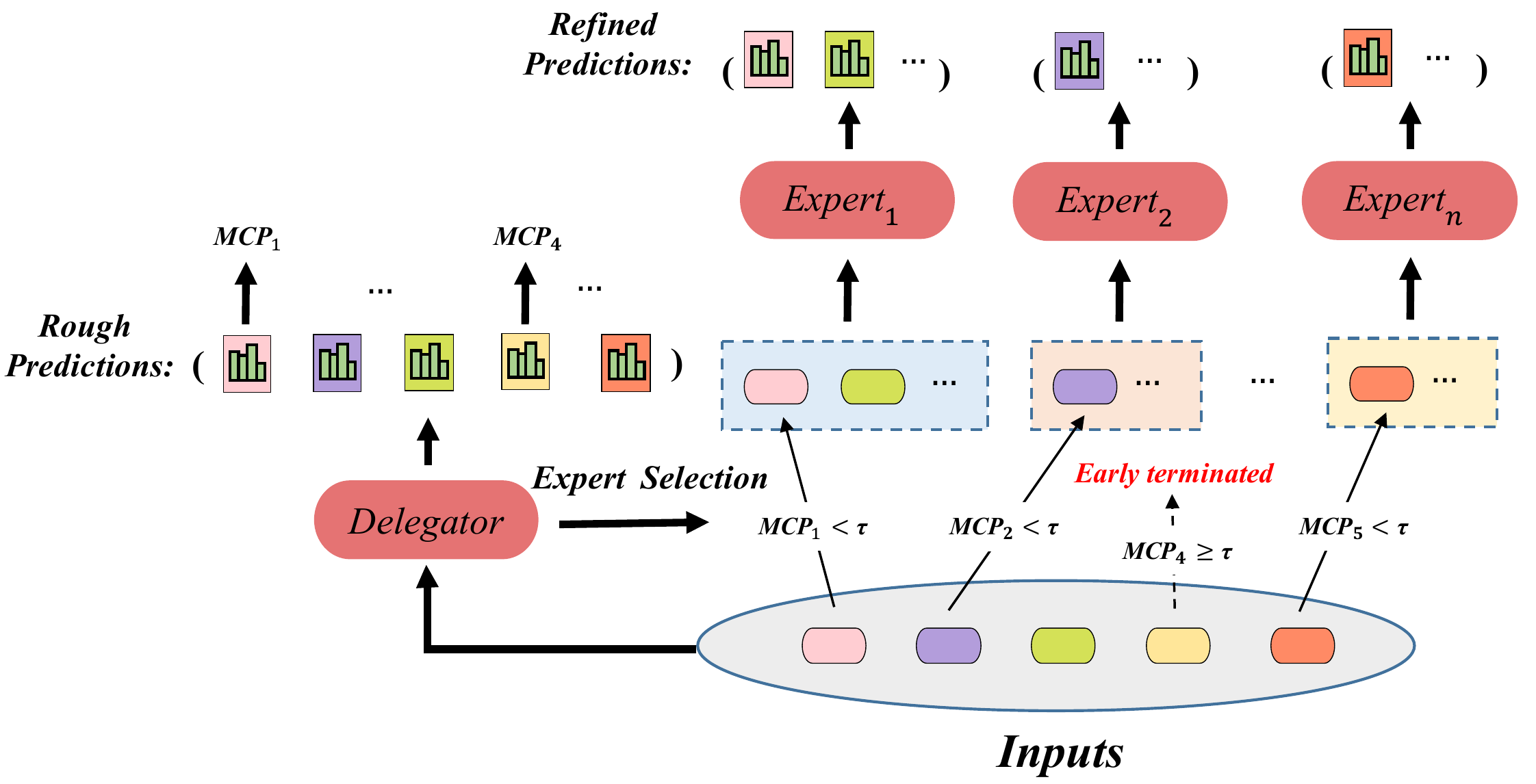}
	\caption{The inference procedure of CoE. MCP is the probability of predicted class. $\tau$ is the threshold for early termination.}
	\label{inference}
\end{figure}

CoE firstly uses delegator to obtain the rough prediction and determine the selected expert for each sample. Afterward, \textit{Maximum Class Probability} (MCP, \citealt{corbiere2019addressing}) of rough prediction is calculated. It is the probability of predicted class. Then, the final recognition results for samples with MCP larger than a given threshold $\tau$ are derived from the rough predictions. Other samples are partitioned into $n$ groups based on which expert is selected. Subsequently, batch processing can be conducted within each group to obtain refined predictions. This procedure is shown in Fig.\ref{inference}. The averaged FLOPs/Instance of CoE ranges from $F_{D} $ to $F_{D}+F_{E}$ by varying $\tau$ from 0 to 1, $F_{D}$ and $F_{E}$ are FLOPs of delegator and experts. Therefore, the value of $\tau$ is directly determined by target FLOPs.

\subsection{Label Generation Module (LGM)}\label{module1}

Since expert selection for CoE is done across models, we can measure the suitability of each expert without bias. Based on this, labels to supervise delegator for expert selection can be obtained. This label-based training method for delegator is more reasonable than the widely-used gate-value-based method~\citep{shazeer2017outrageously, fedus2021switch} which enables the training of routing function by using its predicted gate-value to scale the output of each expert. Next, we firstly introduce how to measure the suitability of each expert, then illustrate how to obtain the selection label.

Model accuracy can be measured by \textit{True Class Probability} (TCP, \citealt{corbiere2019addressing}):
\begin{equation}
\label{tcp_eq}
\text{TCP}_{j,k}=P(Y=y_j|x_j,\text{Expert}_k),
\end{equation}
where, $x_j$ is the j-th sample, $Y$ and $y_j$ are the predicted and true class. {But accuracy is not the only factor for suitability. For example, when models are of different sizes, the larger one usually has a higher TCP. But it may not be more suitable, due to the large inference cost. Given that our concern is not the optimization of network architecture, we can suppose no expert is superior to others (\hypertarget{nsa}{\textit {No Superiority Assumption}, NSA}). Motivated by NSA, we leverage the standardized TCP as the metric for sutability}:
\begin{equation}
\label{suitability}
S_{j,k}=\frac{\text{TCP}_{j,k}-Mean(\text{TCP}_{:,k})}{Std(\text{TCP}_{:,k})},
\end{equation}
where, $\text{Mean}(\text{TCP}_{:,k})$ and $\text{Std}(\text{TCP}_{:,k})$ are mean value and standard deviation for TCPs of $\text{Expert}_k$ on $m$ samples. 

Selection labels can be denoted by a binary matrix $L_{m \times n}$, where each row is a selection label.  According to \hyperlink{nsa}{NSA}, each expert should be assigned the same number of samples, thus the sum of each colum vector of $L_{m \times n}$ should be same, i.e. $\sum_{j}L_{j,k}=\frac{m}{n}$ for $k=1,...,n$. Therefore, $L_{m \times n}$ can be obtatined by maximizing $\sum_{j,k} S_{j,k} * L_{j,k}$:
\begin{equation}
\label{lgm}
\begin{split}
&\min\quad \sum\limits_{j,k} -S_{j,k} * L_{j,k}\\
& \begin{array}{l@{\quad}l@{}l@{\quad}l}
s.t.
&L_{j,k} \in \{0,1\},~~\sum\limits_{k} L_{j,k} = 1,~~\sum\limits_{j} L_{j,k} = \frac{m}{n}\\
\end{array}
\end{split}
\end{equation}

This problem can be modeled as the balanced transportation problem (BTP, \citealt{shore1970transportation}), where each sample is a supply source with a supply of one, each expert is a demand source with a demand of $m/n$. $-S_{j,k}$ is the per-unit transportation cost from the j-th supply source to the k-th demand source. We solve this problem via Vogel approximation method (VAM, \citealt{shore1970transportation}) as introduced in Appendix~\ref{intro_vam}, which is a short-cut approach to invariably obtain a good solution.

\subsection{Weight Generation Module (WGM)}\label{module2}

To maximize the collective capacity of CoE, the dataset needs to be partitioned into portions then each expert only focuses on one portion. This is achieved by WGM which reweights the losses of experts. The partition can be indicated by an assignment matrix $A_{m\times n}$, with one-hot row vectors. $A_{j,k} = 1$ means the j-th sample $x_j$ is assigned to the k-th expert, thus the loss weight for $\text{Expert}_k$ gets larger than other experts on $x_j$.

\begin{figure}[]
	\centering
	\begin{subfigure}{1.0\linewidth}
		\centering
		\includegraphics[width=1.0\linewidth]{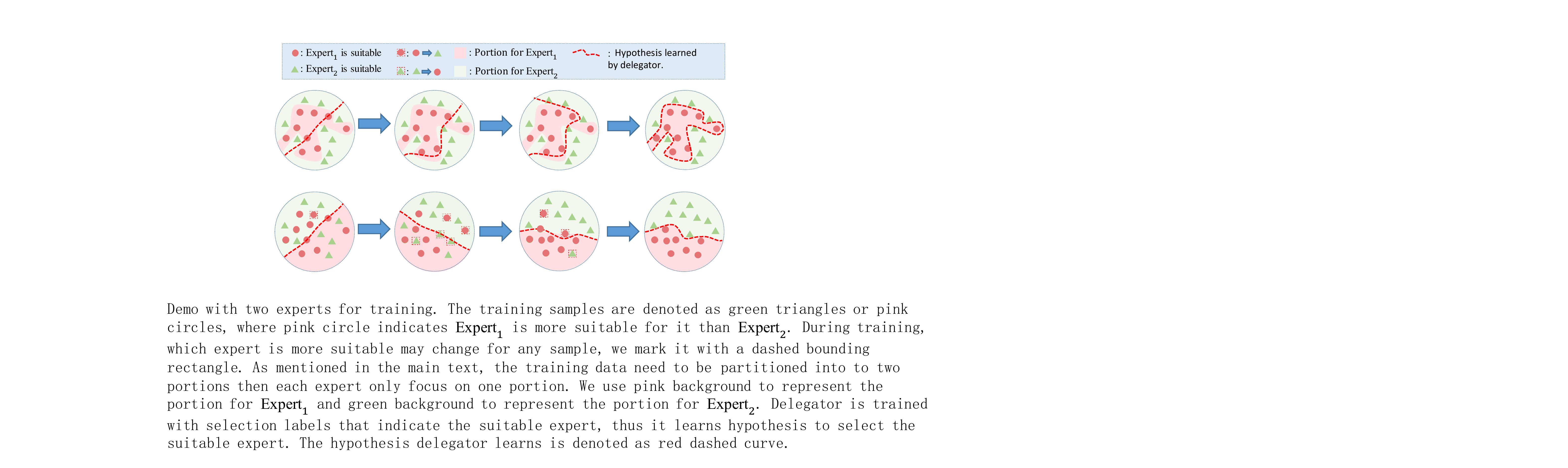}
		\caption{Partition the training data based on expert suitability, which makes delegator overfit to irregular selection labels.}
		\label{partition_A}
	\end{subfigure}
	\begin{subfigure}{1.0\linewidth}
		\centering
		\includegraphics[width=1.0\linewidth]{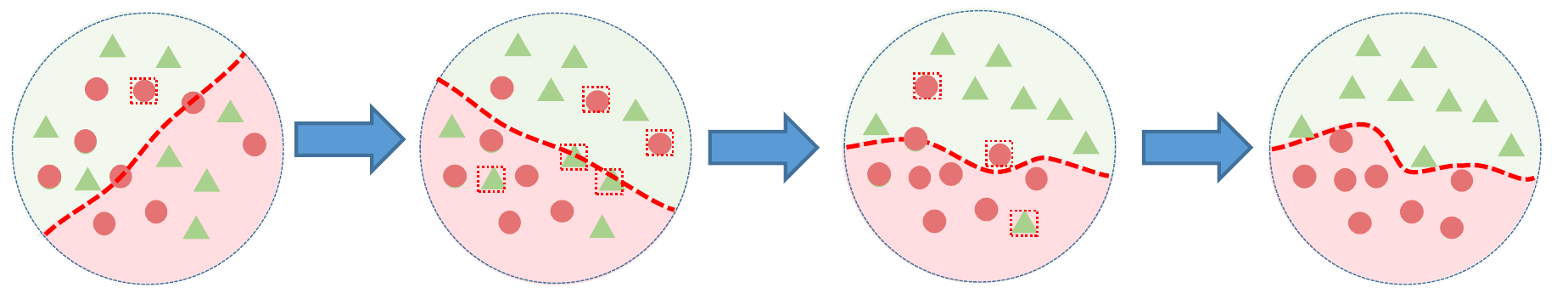}
		\caption{Partition the training data based on delegator, which fulfills the co-adaptation of experts and delegator, avoiding the overfitting problem of delegator.}
		\label{partition_B}
	\end{subfigure} 
	\caption{Demo with two experts of the training process. The training samples are denoted as green triangles or pink circles, where pink circle indicates $\text{Expert}_1$ is more suitable for it than $\text{Expert}_2$. During training, which expert is more suitable may change, we mark it with a dashed bounding square. As mentioned, the training data need to be partitioned into two portions then each expert only focuses on one portion. We use pink background to represent the portion for $\text{Expert}_1$ and the green one for $\text{Expert}_2$. Delegator learns hypothesis to select expert, it is denoted as red dashed curve.}
\end{figure}

A naive partition can be based on expert suitability, namely, partitioning the training data with selection labels $L_{m\times n}$. However, it results in a poor generalization to delegator as shown in Fig.\ref{partition_A}. Assuming $\text{Expert}_k$ is suitable on a sample $x_j$, thus $A_{j,k}=L_{j,k}=1$. Due to $A_{j,k}=1$, the loss weight for $\text{Expert}_k$ gets larger than other experts on $x_j$, making $\text{Expert}_k$ more suitable in return. Therefore selection labels cannot be updated and the irregularity for them caused by random initialization will be preserved. Consequently, delegator gradually overfits to irregular labels, yielding a poor generalization. This is also verified in Section~\ref{ablation_for_coe_components}, where the expert-suitability-based partition results in a terrible performance for CoE$^{WGM^\star}$.

Since networks learn gradually more complex hypotheses during training~\citep{pmlr-v70-arpit17a}, delegator tends to learn generalizable patterns first. Therefore, the partition can be based on generalizable patterns with delegator as the bridge, namely, partition based on the output of delegator. In this way, selection labels get more regular in return due to the reweighting of expert losses. As shown in Fig.\ref{partition_B}, delegator avoids overfitting to the irregular labels.

Delegator outputs a probability matrix $P_{m\times n} \in R^{m\times n}$, whose element $P_{j,k}\in[0,1]$ represents the probability of selecting the k-th expert on the j-th sample. As analyzed above, it is better to partition the training data based on $P_{m\times n}$, thus  $A_{m\times n}$ is obtained by maximizing $\sum_{j,k}P_{j,k} * A_{j,k}$. Moreover, according to \hyperlink{nsa}{NSA}, the number of samples assigned to each expert should be same, i.e. $\sum_{j}A_{j,k} =m/n$. Thus, $A_{m\times n}$ is optimized by:

\begin{equation}
\label{wgm}
\begin{split}
&\min\quad \sum\limits_{j,k} -P_{j,k} * A_{j,k}\\
& \begin{array}{l@{\quad}l@{}l@{\quad}l}
s.t.
&A_{j,k} \in \{0,1\},~~
\sum\limits_{k} A_{j,k} = 1,~~
\sum\limits_{j} A_{j,k} = \frac{m}{n}\\
\end{array}
\end{split}
\end{equation}

This problem can also be modeled as BTP, and solved via VAM as described in section~\ref{module1}.
In the early training stage, the models in CoE are underfitted. Thus we cannot trust $A_{m\times n}$ and need to make the gap between loss weights for different experts smaller to warm up. We achieve this by smoothing $A_{m\times n}$ to $\overline{A}_{m\times n}$ with Eq.\ref{smooth_A}, where $\alpha$ grows linearly from 0.2 to 0.8 with the training going on, 

\begin{equation}
\label{smooth_A}
\begin{split}
\overline{A}_{j,k}=\left\{\begin{array}{lc}
\alpha + \frac{1-\alpha}{n}, \qquad if\ A_{j,k}=1\\
\frac{1-\alpha}{n},\qquad \qquad if\ A_{j,k}=0\end{array}\right. .
\end{split}
\end{equation}

Finally, the output of WGM (i.e. $W_{m \times n}$) is obtained by normalizing $\overline{A}_{m\times n}$ with the coefficient $\mathcal{Z}=\sum\limits_{j} \overline{A}_{j,k}=\frac{m}{n}$: 

\begin{equation}
\label{norm_A}
W_{j,k}=\frac{\overline{A}_{j,k}}{\mathcal{Z}}.
\end{equation}

\subsection{Training Details}

The training framework of CoE is shown in Fig.\ref{framework}, which consists of three major losses: $\mathcal{L}_{P}$, $\mathcal{L}_{T}$ and $\mathcal{L}_{S}$. 

$\mathcal{L}_P$ is the cross-entropy loss for the rough prediction of delegator. To avoid the repeated training of delegator, we use $\mathcal{L}_P$ to train the feature extractor and task predictor (Fig.\ref{generalist}) first of all. Then these two modules are fixed, only expert selector and experts are jointly optimized with $\mathcal{L}_{Total} = \eta * \mathcal{L}_S +\mathcal{L}_T$, $\eta$ is set as 0.8 in this paper. 

$\mathcal{L}_{S}$ is used to optimize the expert selector. Based on the selection label $L_{j,:}$, we can get the cross-entropy loss $\mathcal{L}^j_{S}$ for the j-th sample. Because the final recognition result of CoE is not always sensitive to expert selection, $\{\mathcal{L}^j_{S}|j=1,\dots,m\}$ should be attached various importance. For example, when experts have similar suitabilities (Eq.\ref{suitability}) on the j-th sample, expert selection will have little influence to final performance of CoE, therefore the weight for $\mathcal{L}^j_{S}$ gets smaller. Considering the similarity of suitabilities can be measured by the standard deviation $\text{Std}(S_{j,:})$, we set the loss weight for $\mathcal{L}^j_{S}$ as $v_j=\frac{Std(S_{j,:})}{\sum_{i}Std(S_{i,:})}$. Finally, 

\begin{equation}
\mathcal{L}_{S}=\sum\limits_{j}v_j*\mathcal{L}^j_{S}.
\label{selection_reweight}
\end{equation}

$\mathcal{L}_T$ is used to optimize the experts. Based on the class labels of m samples, we can get $m\times n$ cross-entropy losses $\{{\mathcal{L}_T^{j,k}}|j=1,\dots,m;k=1,\dots,n\}$, where $\mathcal{L}_T^{j,k}$ is the cross-entropy loss for the $k$-th expert on the $j$-th sample. Then $\mathcal{L}_T$ is obtained by the weighted sum of ${\mathcal{L}_T^{j,k}}$ with weights $W_{j,k}$ output by WGM: 

\begin{equation}
\mathcal{L}_{T} = \sum\limits_{j,k}W_{j,k} * {\mathcal{L}_T^{j,k}}.
\end{equation}

We use either four or sixteen experts in this paper. When using sixteen experts, we decompose the task into four subtasks, each of which involves four experts as described in Appendix~\ref{strategy}. This reduces the memory cost for training.

\section{Experiments}
We conduct the main experiments on ImageNet classification task. After comparing with some popular efficient models, we verify the superiority of CoE over the existing model collaboration methods: model ensemble and category-based model selection. Afterward, the effectiveness of training strategy is analyzed by the comparison with the widely-used gate-value-based training method and the elaborated ablations.  Moreover, we try to generalize CoE to the translation task and re-evaluate CoE using Reassessed Labels (ReaL)~\citep{DBLP:journals/corr/abs-2006-07159}. Finally, we try to analyze the reasonability of learned expert selection patterns. Statistics on referenced baselines in section~\ref{main_results}\&\ref{speed_on_hardware} are directly cited from original papers, others are implemented with the following setting unless otherwise stated.

\subsection{Implementation Details}
\label{details}
We conduct experiments with two settings: \textbf{CoE-Small} and\textbf{ CoE-Large}. For CoE-Small, we take TinyNet-E~\citep{DBLP:journals/corr/abs-2010-14819} with 24M FLOPs as the feature extractor of delegator by removing the last fully connected layer. We use OFA-110~\citep{cai2019once} with 110M FLOPs as the expert. For CoE-Large, MobileNetV3-Small~\citep{howard2019searching} with 56M FLOPs is adopted to construct the delegator by analogy. We use OFA-230 with 230M FLOPs as the expert. We have also tried to introduce CondConv~\citep{zhang2020dynet} and PWLU activation fuction~\citep{DBLP:journals/corr/abs-2104-03693} to achieve the extreme performance. More details are illustrated in Appendix~\ref{detail_for_ImageNet}.

\subsection{Results and Analysis}

\subsubsection{Accuracy and Computation Cost}\label{main_results}

\begin{figure}[]
	\centering
	\includegraphics[width=1.0\linewidth]{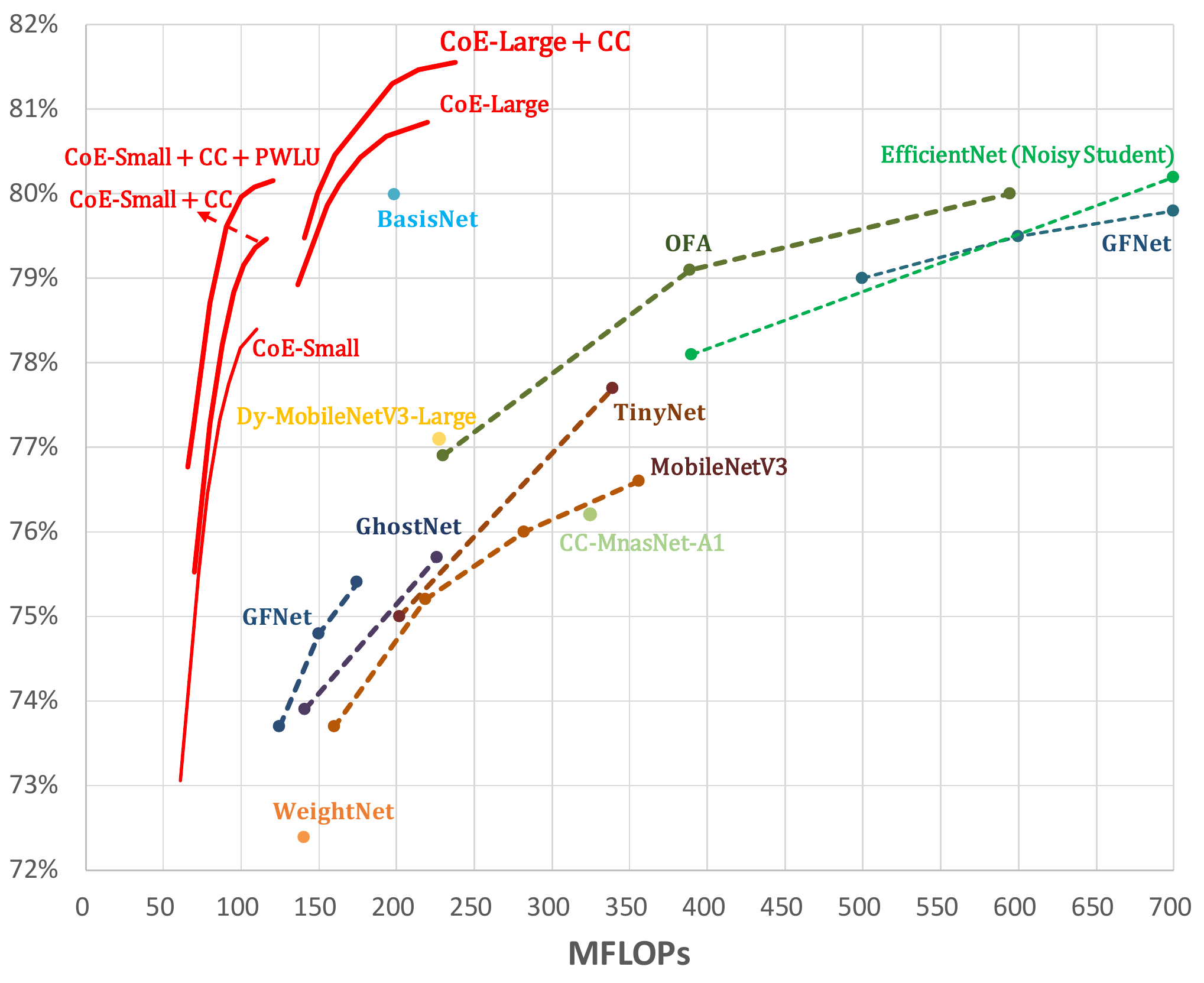}
	\vspace{-10pt}
	\caption{Top-1 Accuracy v.s. FLOPs for CoE on ImageNet. ``CC'' means CondConv and 'PWLU' is an activation function.}
	\label{accuracyv2}
\end{figure}

By varying the threshold $\tau$ of early termination, the
accuracy curves for CoE are obtained. We show them in Fig.~\ref{accuracyv2}, then pick out a point from each curve to compare with some efficient models in Table~\ref{main_table}. CoE achieves 78.2\% and 80.7\% accuracy with 16 experts and the averaged FLOPs/Instance are 100M and 194M respectively. Compared with OFA, CoE reduces the FLOPs from 230M to 100M and from 595M to 194M, with better top-1 accuracy. Though dynamic networks like GFNet, CondConv and BasisNet are more efficient than traditional networks, CoE still has significantly higher accuracy with smaller FLOPs. Compared with these approaches, CoE improves the accuracy by 2.2/2.4/0.7\% respectively. When combined with CondConv, CoE achieves 79.2\% and 81.5\% accuracy with only 102M and 214M FLOPs, indicating that CoE is complementary to dynamic networks like CondConv. On the contrary, as CondConv and BasisNet share similar essence, namely using a group of basis to dynamically synthesize the input-dependent convolution kernel, the combination of them only arouses little collaborative benefit with the accuracy as 80.5\%. Moreover, CoE achieves 80.0\% accuracy with 100M FLOPs for the first time by further using PWLU.

\begin{table}[]
\footnotesize
\makeatletter\def\@captype{table}\makeatother\caption{Compare CoE with some efficient models on ImageNet.}
\label{main_table}
\centering
\begin{tabular}{lcc}
	\toprule
	\multicolumn{1}{c}{Method}               &FLOPs & TOP-1 Acc \\
	\midrule
	WeightNet~\citeyearpar{DBLP:journals/corr/abs-2007-11823}                             &141M  & 72.4\%    \\
	DS-MBNet-M$^{\dagger \ddagger}$\citeyearpar{Li_2021_CVPR}                            & 319M  & 72.8\%    \\
	GhostNet 1.0x~\citeyearpar{DBLP:journals/corr/abs-1911-11907}                            &141M  & 73.9\%    \\
	MobileNetV3-Large~\citeyearpar{howard2019searching}                & 219M  & 75.2\%    \\
	OFA-230~\citeyearpar{howard2019searching}                         &230M  & 76.9\%    \\
	TinyNet-A~\citeyearpar{DBLP:journals/corr/abs-2010-14819}                         & 339M  & 77.7\%    \\
	CondConv-EfficientNet-B0~\citeyearpar{yang2019condconv}          & 413M  & 78.3\%    \\
	GFNet~\citeyearpar{wang2020glance}                            & 400M  & 78.5\%    \\
	\midrule
	CoE-Small                                     & 100M   & 78.2\%    \\
	CoE-Small +   CondConv                         & 102M   & 79.2\%    \\
	CoE-Small +   CondConv + PWLU                         & 100M   & 80.0\%    \\
	\midrule
	BasisNet~\citeyearpar{zhang2021basisnet}                        & 198M  & 80.0\%    \\
	OFA-595~\citeyearpar{howard2019searching}                          & 595M  & 80.0\%    \\
	EfficientNet-B2~\citeyearpar{tan2019efficientnet}                 & 1.0B  & 80.1\%    \\
	EfficientNet-B1(Noisy Student)~\citeyearpar{xie2020self} & 700M  & 80.2\%    \\
	BasisNet~\citeyearpar{zhang2021basisnet}                          & 290M  & 80.3\%    \\
	FBNetV3-C~\citeyearpar{dai2020fbnetv3}                       & 557M  & 80.5\%    \\
	BasisNet + CondConv~\citeyearpar{yang2019condconv}               & 308M  & 80.5\%    \\
	\midrule
	CoE-Large                                     & 194M  & 80.7\%    \\
	CoE-Large + CondConv                            & 214M  & 81.5\%   \\
	\bottomrule
\end{tabular}
\end{table}

\subsubsection{Inference Speed and Memory Cost}
\label{speed_on_hardware}
The core advantage of CoE is the inference efficiency. To verify this advantage, we analyze the inference latency on hardware. The experiments are conducted on CPU platform (Intel(R) Xeon(R) CPU E5-2699 v4 @ 2.20GHz) with PyTorch version as 1.8.0. We report the averaged latency on the ImageNet validation set in Table~\ref{latency}. We notice the discrepancy between FLOPs and real speed. For example, OFA-230 has 1.6x FLOPs compared with GhostNet 1.0x, but the speed is 1.2x faster. Moreover, this discrepancy can be enlarged by CondConv. CondConv-EfficientNet-B0 has similar FLOPs with the original EfficientNet-B0, but the speed is 1.7x slower. BasisNet synthesizes the dynamic parameters all at once, rather than the ``layer by layer'' fashion like CondConv, thus is more efficient. However, it still needs to load a large number of parameters for this synthesis, which brings a large MAC. This is why CoE (16 experts) can reduce 14.09\% latency than BasisNet even when the mini-batch size is one. Finally yet importantly, BasisNet and CondConv do not support batch processing, while CoE (16 experts) can take advantage of it to further achieve a 3.1/6.1x speedup compared with them. {We analyze the memory cost from two perspectives: the number of parameters and MAC. As can be seen from Table~\ref{latency}, the accuracy of CoE-Large (4 experts) is no worse than BasisNet and CondConv-EfficientNet-B0 when using similar parameters. Besides, the averaged MAC/Instance of CoE is much smaller than theirs.

\begin{table*}
	\centering
	\footnotesize
	\caption{CPU latency and memory cost for different methods.}
	\label{latency}
	\begin{tabular}{l|c|c|cccc}
		\toprule
		\multicolumn{1}{c|}{\multirow{2}{*}{Models}} & \multicolumn{2}{c|}{CPU   Latency/Instance (ms)} & \multirow{2}{*}{FLOPs} &\multirow{2}{*}{{MAC}} &\multirow{2}{*}{{Params}} & \multirow{2}{*}{Accuracy} \\\cline{2-3}
		\multicolumn{1}{c|}{}                        & Batchsize=1           & Batchsize=64           &                        &                           \\
		\cline{1-7}
		MobileNetV3-Small                              & 14.77                 & 4.18                  & 56M&{2.5M}&         {2.5M}          & 67.4\%                    \\
		GhostNet 1.0x                               & 39.91                 & 16.50                 & 141M & {5.2M} &      {5.2M}          & 73.9\%                    \\
		TinyNet-B                                   & 34.58                 & 19.44                  & 202M&{3.7M} &   {3.7M}               & 75.0\%                    \\
		{MobileNetV3-Large}                               &  {31.55}                 & {18.43}                  & {219M }&{5.4M}&               {5.4M}   & {75.2\%}                   \\
		
		GhostNet 1.3x                               &  43.94                 & 29.70                  & 226M& {7.3M}  &  {7.3M}              & 75.7\%                    \\
		OFA-230                                     & 33.52                 & 15.21                  & 230M&    {5.8M}  &   {5.8M}          & 76.9\%                    \\
		EfficientNet-B0                                    & 49.12                 & 35.21                      & 391M &{5.3M}&    {5.3M}             & 77.2\%                    \\
		TinyNet-A                                   & 45.76                 & 23.71                  & 339M&   {5.1M}     &  {5.1M}         & 77.7\%                    \\
		\cline{1-7}
		CondConv-EfficientNet-B0                                    & 81.81                 & -                      & 413M&             {24.0M}     &{24.0M} & 78.3\%                    \\
		BasisNet                                    & 40.61                 & -                      & 198M &   {24.9M}    &    {24.9M}       & 80.0\%                    \\
		{CoE-Large (4 experts)}                                        & {38.67}
		& {15.02}                &{ 220M }&     {6.6M}     &    {25.7M}    & {79.9\%}\\ 
		CoE-Large (16 experts)                                    & 34.89                 & 13.30                  & 194M &     {6.0M}     &    {95.3M}    & 80.7\%\\\bottomrule                 
	\end{tabular}
\end{table*}

\subsubsection{Analysis of the training cost}

To achieve superior performance with fewer inference FLOPs and latency, CoE may consume more training time. For example, the applying of CoE (4 experts) on OFA-230 improves the accuracy from 78.0\% to 79.9\%, but at the expense of a 2.2x training cost. To verify whether the improvement still exists with similar training cost, we get a series of accuracies by varying the number of training epochs as shown in Fig.\ref{training_cost}, where 32 GPUs (Tesla-V100-PCIe-16GB) are used. It is seen that CoE boosts the performance from 78.3\% to 79.9\% even when the training cost is similar.

\begin{figure}[tbh]
	\centering
	\includegraphics[width=0.9\linewidth]{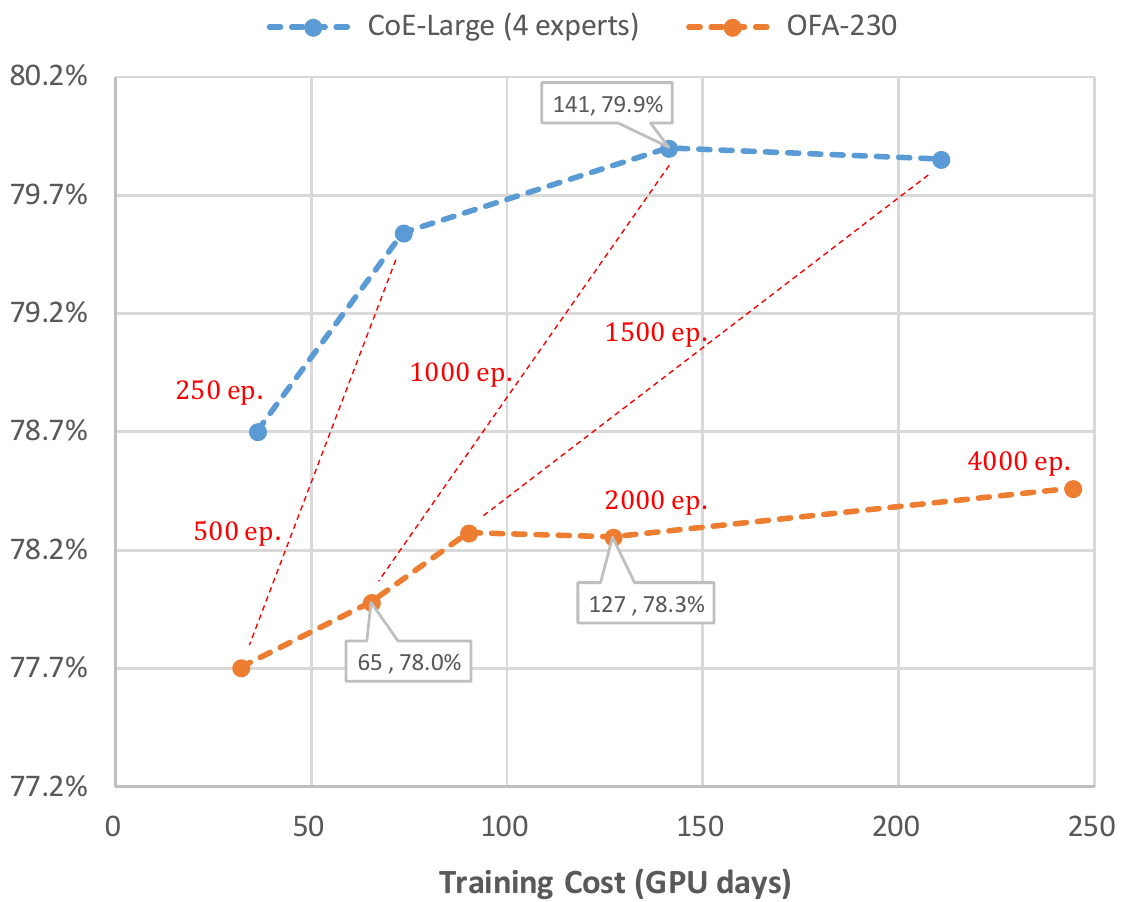}
	\caption{Accuracy v.s. training cost for OFA-230 and CoE-Large (4 experts). ``xx ep.'' means the number of training epochs is ``xx''.}
	\label{training_cost}
\end{figure}

\subsection{Comparison with Existing Approaches}

\subsubsection{Comparison with Model Ensemble}

We train four OFA-230 models with different initialization seeds as shown in Table~\ref{ensemble}. The random initialization usually causes minor variety in accuracy for ImageNet classification but leads models to fall into different local minima, yielding the diversity of output. This diversity enables the ensembled models to achieve an improvement of 1.6\% in accuracy. We adopt the naive ensemble here, i.e. averaging the output of each model. As shown in \citet{wen2020batchensemble, Chen_2019_CVPR}, though simple, naive ensemble has a competitive performance in terms of accuracy. However, CoE still achieves 1.1\% higher accuracy than it. Recently proposed ensemble methods~\citep{havasi2020training} mainly focus on reducing the computation cost with a little drop of accuracy than naive ensemble, but the computation cost is always larger than the one of a base model. By contrast, CoE reduces the FLOPs to 0.84$\times$ of the base model, indicating the superiority of CoE in terms of FLOPs as well.

\begin{table}
\caption{Comparison with model ensembe.}
\label{ensemble}
\centering
\footnotesize
\begin{tabular}{c|c|c|c}
	\toprule
	\multicolumn{2}{c|}{Method}                & FLOPs & Acc. \\
	\toprule
	\multirow{5}{*}{OFA-230} & Seed1           & 230M  & 78.1\%     \\
	& Seed2           & 230M  & 78.0\%     \\
	& Seed3           & 230M  & 78.1\%     \\
	& Seed4           & 230M  & 78.0\%     \\ \cline{2-4} 
	& Ensemble        & 920M  & 79.6\%     \\ \hline
	\multirow{2}{*}{CoE-Large}    & 4 \ Experts    & 194M  & 79.8\%     \\
	& 16   Experts & 194M  & 80.7\%\\
	\bottomrule  
\end{tabular}
\end{table}

\subsubsection{Comparison with Category-Based Method} 

HD-CNN and HydraNets select branches based on the category. Despite their methods are originally designed to select a specific block, we apply them to the model level. To select expert based on category, the categories should be partitioned into n groups, where n is the number of experts. We try two schemes: random partition and clustering-based partition. Then, expert can be selected according to the
rough prediction of delegator. During the training procedure, we also reweight losses of each expert based on the assignment matrix $A_{m\times n}$ with Eq.\ref{smooth_A}\&\ref{norm_A}. Here, $A_{m\times n}$ is obtained directly based on the rough prediction. The results with \textbf{4 experts} are shown in Table~\ref{cc_approaches}, demonstrating a better collaboration pattern is learned by CoE.

\begin{table}
	\footnotesize
	\centering
	\caption{Comparison with category-based selection. ``RP'' and ``CBP'' means Random Partition and Clustering-Based Partition.}
	\label{cc_approaches}
	\begin{tabular}{c|c|c|c}
		\toprule
		\multicolumn{2}{c|}{Method}                                                  & FLOPs                 & Top-1 Acc.              \\
		\toprule
		\multirow{2}{*}{Category-Based} & \multirow{1}{*}{RP}   & \multirow{1}{*}{220M} & \multirow{1}{*}{78.3\%}\\
		& \multirow{1}{*}{CBP} & \multirow{1}{*}{220M} & \multirow{1}{*}{77.5\%} \\
		\midrule
		\multirow{1}{*}{CoE-Large}
		&-            & 220M                  & 79.9\%    \\ \bottomrule            
	\end{tabular}
\end{table}

\subsubsection{Comparison with Gate-Value-Based Training Method} 

MoE and Switch Transformer adopt the gate-value-based training method for routing function. They enable the training of router by using its predicted gate-value to scale the output of each expert.  This optimization manner is heuristic while CoE trains the delegator more reasonably. Since expert selection for CoE is done across models, we can measure the suitability of each expert without bias. Thereafter, selection labels can be obtained to supervise delegator. Despite the gate-value-based method is originally designed to select a specific layer, we apply it to the model level. We compare with both the soft gate-value and the hard gate-value. For hard gate-value, it is a one-hot vector generated by replacing the softmax in expert selector (Fig.\ref{generalist}) with gumbel softmax~\citep{2017Categorical}. The results with \textbf{4 experts} are shown in Table~\ref{gate-value}, where CoE achieves better performance, indicating the effectiveness of our training method.

\begin{table}[tbh]
	\footnotesize
	\centering
	\caption{Comparison with gate-value-based training method.}
	\label{gate-value}
	\begin{tabular}{c|c|c|c}
		\toprule
		\multicolumn{2}{c|}{Method}                                                  & FLOPs                 & Top-1 Acc.              \\
		\toprule
		\multirow{2}{*}{Gate-Value-Based} & \multirow{1}{*}{Soft Gate-Value}   & \multirow{1}{*}{220M} & \multirow{1}{*}{78.7\%}\\
		& \multirow{1}{*}{Hard Gate-Value} & \multirow{1}{*}{220M} & \multirow{1}{*}{78.9\%} \\
		\midrule
		\multirow{1}{*}{CoE-Large}
		&-            & 220M                  & 79.9\%    \\ \bottomrule            
	\end{tabular}
\end{table}

\subsection{Ablation Studies for CoE}
\label{ablation_for_coe_components}

We have conducted elaborated ablations, including ablations for each element of our proposed training method, ablations for the training tricks, ablations for expert number and early termination. We mainly introduce the ablations for our proposed training method here, others are discussed in Appendix~\ref{ablation}. 

CoE consists of 2 major components: LGM and WGM. Apart from directly removing one component, we also try to alter some elements inside them. We propose several modified versions of CoE for ablation as below:

\begin{itemize}
	\setlength{\itemsep}{0pt}
	\item CoE$^{LGM}$: Remove LGM from CoE. Thus, CoE collapses to a single expert with delegator to trigger the early termination. 
	\item CoE$^{LGM^\star}$: Abandon the refining of suitability criterion (Eq.\ref{suitability}) and remove the ``$\sum_{j}L_{j,k} =m/n$''constraint in Eq.\ref{lgm}. So that $L_{m\times n}$ neglects the \textit{No Superiority Assumption} (\hyperlink{nsa}{NSA}).
	\item CoE$^{WGM}$: Remove WGM from CoE. Thus, losses of experts have identical weights for each sample.
	\item CoE$^{WGM^\star}$: WGM partitions the training data based on expert suitability, thus the assignment matrix $A_{m\times n}$ in WGM equals to the output matrix $L_{m\times n}$ of LGM.
	\item CoE$^{WGM^\circ}$: Remove the ``$\sum_{j}A_{j,k} =m/n$'' constraint in Eq.\ref{wgm} and abandon the smoothing of $A_{m\times n}$ (Eq.\ref{smooth_A}). Thus WGM neglects the \hyperlink{nsa}{NSA}.
	\item CoE$^{WGM^\bullet}$: Abandon the progressive sharpening for $A_{m\times n}$ in WGM. Specifically, set $\alpha$ in Eq.\ref{smooth_A} as a constant 0.8, instead of linearly increasing it.
	\item CoE$^{SR}$: Abandon the reweighting for losses of expert selection ($\mathcal{L}_S^j$), namely set $v_j$ in Eq.\ref{selection_reweight} as a constant $\frac{1}{m}$.
\end{itemize}

The results for those CoE versions with the CoE-Large setting and 4 experts are shown in Table~\ref{component_ablation}, which demonstrates the significance of each element of the training method.

\begin{table}[tbh]
\centering
	\footnotesize
	\caption{{Ablations for each component of the training method.}}
	\label{component_ablation}
	\begin{tabular}{cccc}
		\toprule
		Method                         & Experts & FLOPs & Acc.    \\
		\midrule
		CoE-Large                        & 4           & 220M  & 79.9\%       \\
		\midrule
		CoE-Large$^{LGM}$~~~~                        & 4           & 220M  & 78.0\%       \\
		CoE-Large$^{LGM^\star}$~~                       & 4           & 220M  & 79.4\%       \\
		CoE-Large$^{WGM}$~~~                      & 4           & 220M  & 78.1\%       \\
		CoE-Large$^{WGM^\star}$                      & 4           & 220M  & 77.0\%       \\
		CoE-Large$^{WGM^\circ}$                      & 4           & 220M  & 79.2\%       \\
		CoE-Large$^{WGM^\bullet}$                        & 4           & 220M  & 79.4\%       \\
		CoE-Large$^{SR}$~~~~                        & 4           & 220M  & 79.5\%       \\
		\bottomrule
\end{tabular}
\end{table}

\subsection{Analysis of the Generalization}

To verify the generalizability, we conduct two extra experiments: {\textit {\textbf{generalizing CoE to translation task}} and using Reassessed Labels (ReaL)~\citep{DBLP:journals/corr/abs-2006-07159} to {\textit {\textbf{re-evaluate CoE}}. We mainly introduce the first one here, another one are discussed in Appendix~\ref{re_evaluation}.
		
To generalize CoE to translation task, we build a CoE-Transformer model based on Transformer (base model)~\citep{attentionis}. CoE-Transformer has four decoders, given a sentence, one decoder will be selected to decode the features extracted by encoder. To select the decoder, an extra constant token $\phi$ is added at the beginning of each sentence, whose feature extracted by encoder is input to the expert selector (Fig.\ref{generalist}) for expert selection. During training, the TCP of a sentence is obtained by averaging the TCPs of each token. The architecture of CoE-Transformer are shown in Fig.~\ref{coetransformer}.

\begin{figure}[]
	\centering
	\includegraphics[width=0.8\linewidth]{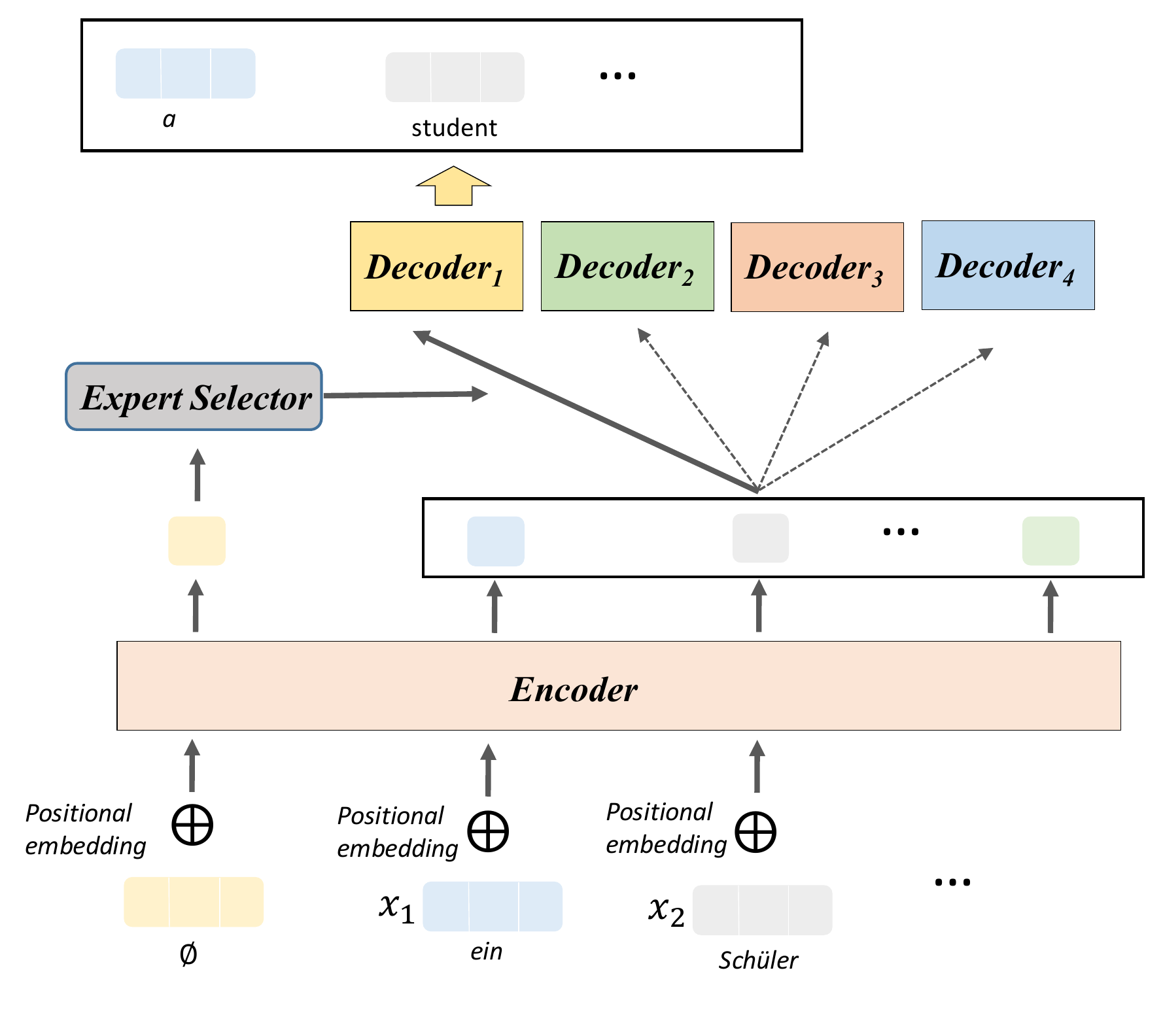}
	\caption{The architecture of CoE-Transformer. $\phi$ is a constant token, whose feature is used for expert selection.}
	\label{coetransformer}
\end{figure}

Following \citep{attentionis,ott2019fairseq}, CoE-Transformer is trained on the standard WMT 2014 English-German dataset. As mentioned, an extra token will be added to this vocabulary. We adopt the same training and evaluating setting as \citep{ott2019fairseq}, more details are shown in Appendix~\ref{trans_details}. From Table~\ref{translation_result} we can see, CoE-Transformer outperforms Transformer (base model) by a large margin and achieves similar performance as Transformer (big) with much less MAC and parameters.

\begin{table}[tbh]
	\footnotesize
	\caption{The BLEU scores on newstest2014 (English-to-German).}
	\label{translation_result}
	\centering
	\begin{tabular}{cccc}
		\toprule
		Model                         & MAC & Parameters  & BLEU    \\
		\midrule
		Transformer (base)&62.4M & 62.4M  &28.1~\citeyearpar{ott2019fairseq}\\
		\midrule
		Transformer (big)&213.0M & 213.0M  &29.3~\citeyearpar{ott2019fairseq}\\
		\midrule
		CoE-Transformer&62.5M & 138.2M  &29.4\\
		\bottomrule
	\end{tabular}
\end{table}

\subsection{Analysis of the Learned Expert Selection Patterns}

We also conduct experiments to analyze the expert selection patterns of CoE and find them quite reasonable. When experts have different architectures, the delegator tends to assign easy samples to smaller experts and complex samples to heavier experts. When experts share the same architecture, delegator learns the expert selection patterns automatically, it can be based on any property (e.g. whether humans are contained), rather than limited to the category. We introduce the experiment when experts have various architectures here, more details are illustrated in Appendix~\ref{selection_pattern}.

Considering TCP~\citep{corbiere2019addressing} measures the complexity of a given sample if the inference model is fixed, namely, the more complex is the sample, the smaller TCP will be. We can analyze the relationship between sample complexity and expert selection. We take four architectures searched via OFA~\citep{cai2019once} as the experts, i.e. OFA-110, OFA-163, OFA-230 and OFA-595. OFA-xx indicates the FLOPs is xx. The delegator is also MobileNetV3-small as described in section~\ref{details}. We obtain the TCP value for each sample based on the delegator. We count the selection probability for each expert at different TCP values on the validation set. As shown in Fig.\ref{provstcp}, the selection probability for heavier model increases with the input sample getting more complex (with the decrease of TCP). It demonstrates that CoE can learn reasonable expert selection patterns automatically.

\begin{figure}[tbh]
	\centering
	\includegraphics[width=0.9\linewidth]{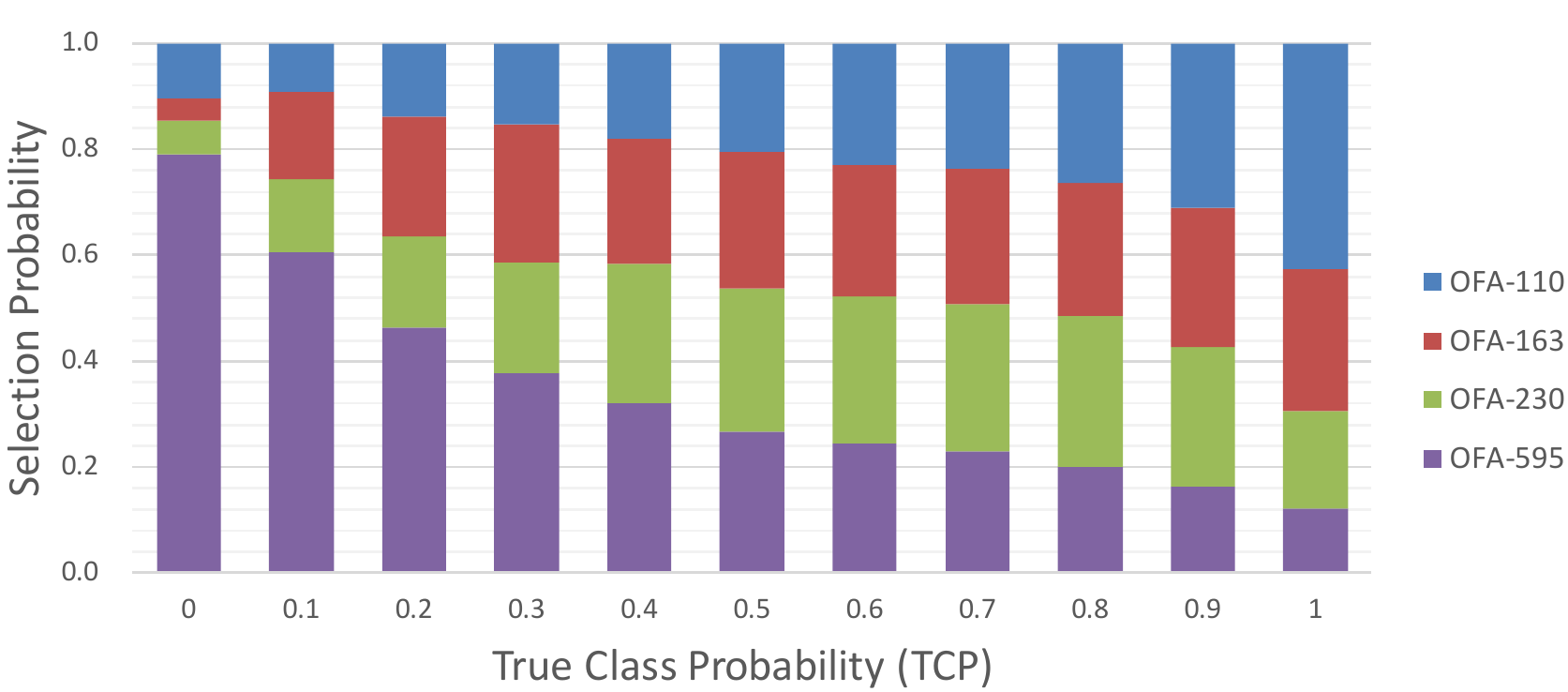}
	\caption{Selection probabilities for each expert at different TCP values. The selection probability for heavier model increases with the input sample getting more complex (with the decrease of TCP)}
	\label{provstcp}
\end{figure}

\section{Conclusion}
We propose a CoE framework to pool together the expertise of multiple networks towards a common aim. Experiments in this paper demonstrate the superiority of CoE on both accuracy and real speed. We also analyze the collaboration patterns and find them have interpretability. In the future, CoE will be extended to the trillion parameters level. Meanwhile, we will try to implement CoE to more tasks and verify its compatibility with quantification and other technologies. Besides, CoE has the potential to solve problems of lifelong learning by updating experts.

\bibliography{paper}
\bibliographystyle{icml2022}

\newpage
\appendix
\onecolumn
\section{Extra Details for Method}
\subsection{Introduction of Vogel Approximation Method (VAM)}
\label{intro_vam}
In weight generation module (WGM) and label generation module (LGM), we need to solve the balanced transportation problem (BTP, \citep{shore1970transportation}) via Vogel approximation method (VAM, \citep{shore1970transportation}). We will introduce it in this section with the number of samples and experts as $m$ and $n$ respectively.

The BTP involved in WGM and LGM has $m$ supply sources, each of which is denoted as $Silo_j$ with a supply of one, as well as $n$ demand sources, each of which is denoted as $Mill_k$ with a demand of $\frac{m}{n}$. $C_{j,k}$ is the per-unit transportation cost from $Silo_j$ to $Mill_k$. Specifically, $C_{j,k} = -P_{j,k}$ in WGM and $C_{j,k} = -S_{j,k}$ in LGM. To make it clear, we illustrate this algorithm with a toy example, where the problem is simplified as Fig.\ref{vam} (a) with $m=4$, $n=2$. In the first step, we calculate the penalty cost $pc_{row_j}$ for each row and $pc_{col_k}$ for each column of the tableau in Fig.\ref{vam} (a). Penalty cost is determined by subtracting the lowest unit cost in the row (column) from the next lowest unit cost. The penalty costs of the respective rows and columns have been marked in red color for clarity in Fig.\ref{vam} (b). Since the third row has the largest penalty cost ($pc_{row_3}$ =11) and $C_{3,1}$ is the lowest unit cost of that row, $Silo_3$ is allocated to $Mill_1$, i.e. $A_{3,:}=[1,0]$ in WGM or $L_{3,:}=[1,0]$ in LGM. Then the corresponding row should be crossed out and the demand of $Mill_1$ should minus one, if this results in a zero demand, the first column will be crossed out as well. After adjusting penalty cost for each row and column, the tableau becomes Fig.\ref{vam} (c), where the changed values are marked in orange. The described procedure will be looped until no rows remained.

Considering the calculation of $pc_{col_k}$ is much more time-consuming compared with $pc_{row_j}$ because $m \gg n$ in WGM and LGM, we modify VAM by only seeking lowest penalty cost among $\{pc_{row_1},...,pc_{row_m}\}$. We find this modification makes VAM more efficient while keeps the superiority of the solution. It is because the mechanics of VAM makes it meaningful to take $pc_{col_k}$ into account only when the demand of  $Mill_k$ is one, which rarely happens. Thus, we adopt this modification to promote efficiency in this paper.

\begin{figure}[tbh]
	\centering
	\includegraphics[width=\linewidth]{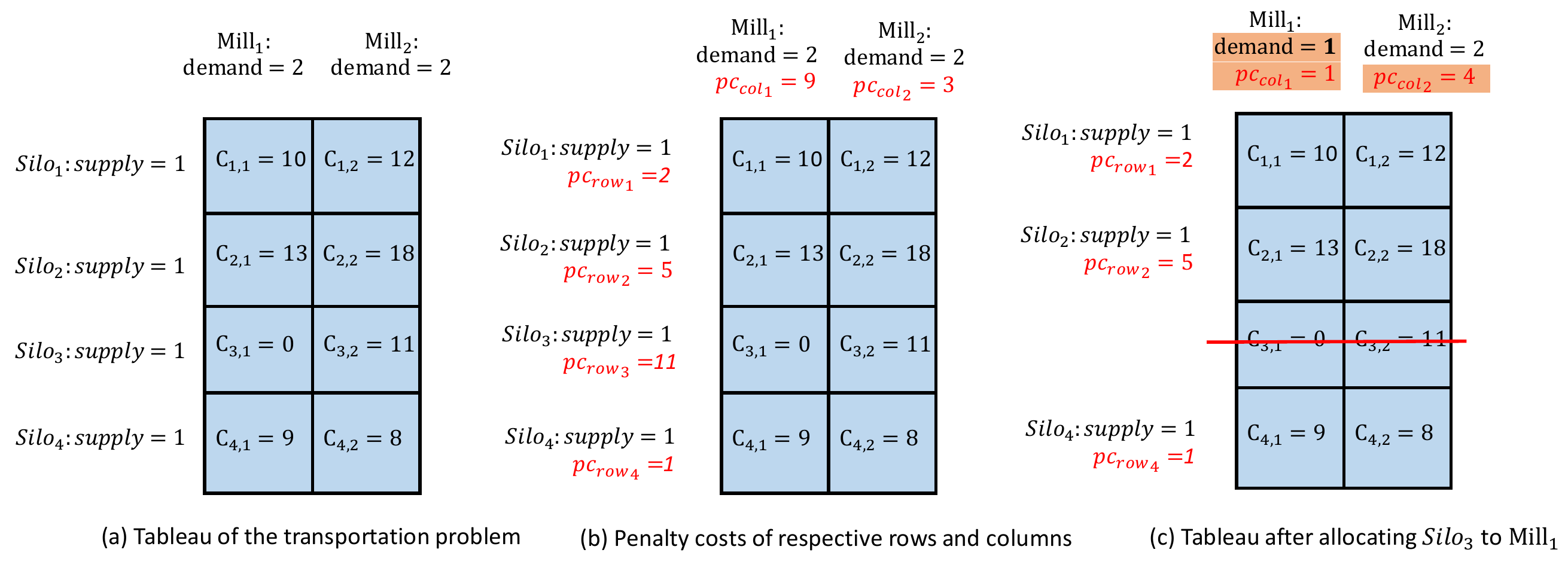}
	\caption{The Vogel approximation method.}
	\label{vam}
\end{figure}

\subsection{A Strategy for Task Decomposition}
\label{strategy}
To fulfill task decomposition, we introduce a new module to delegator, named subtask selector as shown in Fig.\ref{modified_delegator}. The subtask selector is used to allocate the input samples into different subtasks. The expert selector outputs sixteen probabilities, which are partitioned into four groups as well. For each subtask, only one group of probabilities is visible. The experts within each subtask and the corresponding weights of the expert selector are jointly optimized. As for the feature extractor, task predictor, and subtask selector, their weights directly derive from the delegator trained with the setting of four experts and then fixed. During this procedure, the weights of subtask selector derive from the expert selector.

\begin{figure}[]
	\centering
	\includegraphics[width=0.5\linewidth]{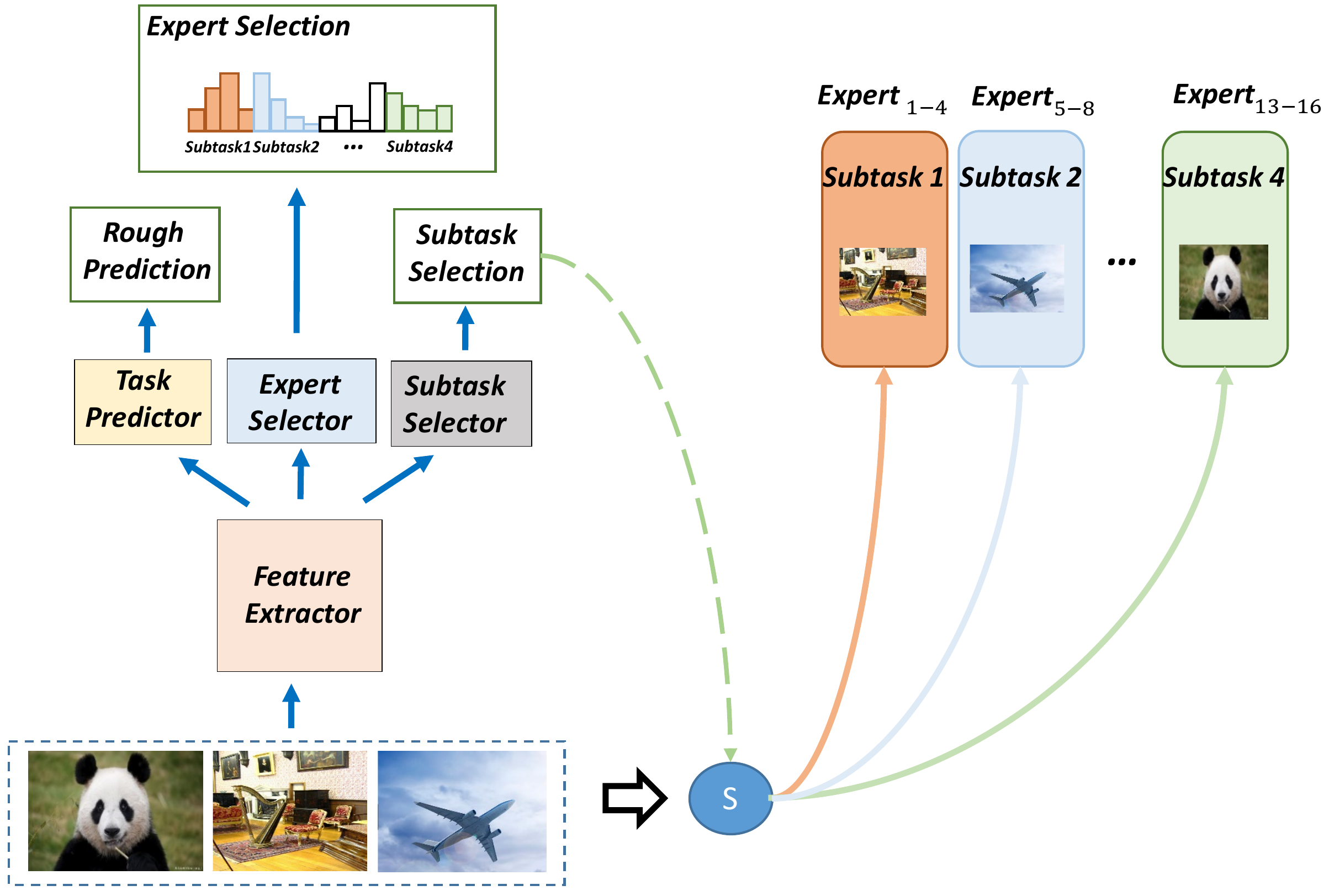}
	\caption{The modified architecture of delegator.}
	\label{modified_delegator}
\end{figure}

\section{Extra details for Experiments}
\subsection{Experiment details for the ImageNet classification task}
\label{detail_for_ImageNet}
To combine with CondConv, we replace the convolutions within each inverted residual block of the experts with CondConv ($expert\_num=4$). To take advantage of PWLU, we replace all activation layers except those that have tiny input feature maps as illustrated in \citet{DBLP:journals/corr/abs-2104-03693}. 
Models are trained using SGD optimizer with 0.9 momentum. We use a mini-batch size of 4096, and a weight decay of 0.00002. Cosine learning rate decay is adopted and the number of training iterations is 313000. We use the augment policy searched by
\citet{cubuk2019autoaugment} as well (fixed auto-augment). Similar as BasisNet, we use knowledge distillation with EfficientNet-B2~\citep{tan2019efficientnet,xie2020self} as the teacher. The learning rate is 0.8/1.6 for CoE-Small/Large and dropout rate is 0.2. The stochastic depth~\citep{huang2016deep} is used except for TinyNet-E with a survival probability of 0.8. \textbf{We think only when the overfitting problem is solved can task accuracy reflect model capacity exactly. Because this paper is concerned with improving the model capacity with limited computation cost, we use knowledge distillation, fixed auto-augment and stochastic depth to overcome the overfitting problem. Nonetheless, we also conduct ablations for them as shown in Appendix~\ref{ablation_for_training_strategy} }

\subsection{Ablation Studies for CoE}
\label{ablation}
\subsubsection{Ablation study for the training tricks}
\label{ablation_for_training_strategy}
Knowledge distillation (KD), auto-augment (AA) and stochastic depth (SD) are widely-used strategies to overcome the overfitting problem. We think only when the overfitting problem is solved can task accuracy reflect model capacity exactly. Because this paper is concerned with improving the model capacity with limited computation cost, we use these strategies. Nonetheless, we conduct ablations for them in this section. We adopt the CoE-Large setting with 4 experts. Results are shown in Table~\ref{training_ablatio}. We find KD extremely important for CoE, it may indicate CoE is easy to be overfitted. In addition, SD decreases the accuracy of CoE. By removing SD, CoE-Large (4 experts) boosts the accuracy from 79.9\% to 80.2\%. Perhaps, it is because SD makes the capacity of delegator and each expert too tiny~\citep{gontijo-lopes2021tradeoffs}.
}

\begin{table}[!tbh]
\centering
\footnotesize
\caption{Ablation study for the training tricks. ``KD'', ``AA'' and ``SD'' denotes knowledge distillation, auto-augment and stochastic depth respectively.}
\label{training_ablatio}
\begin{tabular}{cccccc}
	\toprule
	
	KD&AA&SD&Experts&FLOPs & TOP-1 Acc \\
	\midrule
	$\surd$                           &$\surd$&$\surd$& 4&220M  & \textbf{79.9}\%    \\\midrule
	$\surd$                            &$\surd$&& 4&220M  & \textbf{80.2}\%    \\\midrule
	$\surd$&&$\surd$& 4&220M  & 79.4\%    \\\midrule
	
	&$\surd$&$\surd$& 4&220M  & 76.2\%    \\\midrule
	
	$\surd$&&& 4&220M  & 79.7\%    \\\midrule
	
	&$\surd$&& 4&220M  & \textbf{76.3}\%    \\\midrule
	&&$\surd$& 4&220M  & 75.2\%    \\\midrule
	&&& 4&220M  & 75.1\%    \\
	\bottomrule
\end{tabular}
\end{table}
\subsubsection{Effect of Expert Number}

	We analyze the number of experts in this section, including 1, 4, and 16 experts. The results are shown in Table~\ref{numberofspkls}. Using one expert brings little improvement compared with the original model. When increasing the number of experts, the accuracy becomes 1.9\% better with four experts and 2.9\% better with sixteen experts. It demonstrates CoE can make full use of multiple experts, leading to a large collaborative benefit. What`s more, the accuracy also reaches 79.9\% by combining CondConv with OFA-230. In this manner, CoE can further enhance the accuracy to 80.8/81.5\% with 4/16 experts. 

	\begin{table}[tbh]
		\centering
		\footnotesize
		\caption{Comparison among different number of experts. ``CC'' indicates CondConv.}
		\label{numberofspkls}
		\centering
		\begin{tabular}{p{50pt}<{\centering}p{50pt}<{\centering}p{50pt}<{\centering}p{50pt}<{\centering}}
			\toprule
			Method                         & Experts & FLOPs & Acc.    \\
			\midrule
			OFA-230                        & -           & 230M  & 78.0\%       \\
			\midrule
			\multirow{3}{*}{CoE-Large}          & 1           & 220M  & 78.0\% \\
			& 4           & 220M  & 79.9\% \\
			& 16          & 220M  & 80.9\% \\
			\midrule
			CC-OFA-230                 & -           & 242M  & 79.9\%       \\
			\midrule
			\multirow{3}{*}{CoE-Large + CC} & 1           & 214M  & 79.9\% \\
			& 4           & 214M  & 80.8\% \\
			& 16          & 214M  & 81.5\% \\
			\bottomrule
		\end{tabular}
	\end{table}

\subsubsection{Effect of Early Termination}

The original OFA-230 has 78.0\% top-1 accuracy with 230M FLOPs. We can introduce a MobileNetV3-Small to conduct early termination. By varying the threshold, we get a series of accuracies and FLOPs as shown in Fig.\ref{early_termination}. It can seen that the accuracy becomes 78.0\% with 220M FLOPs. This indicates the computation cost brought by MobileNetV3-Small is eliminated via early termination strategy. Inspired by this, we expect to eliminate the computation cost brought by delegator via early termination as well. It does reduce the computation cost by 60/66M FLOPs, demonstrating the effectiveness of early termination.

\begin{figure}[tbh]
	\centering
	\includegraphics[width=0.7\linewidth]{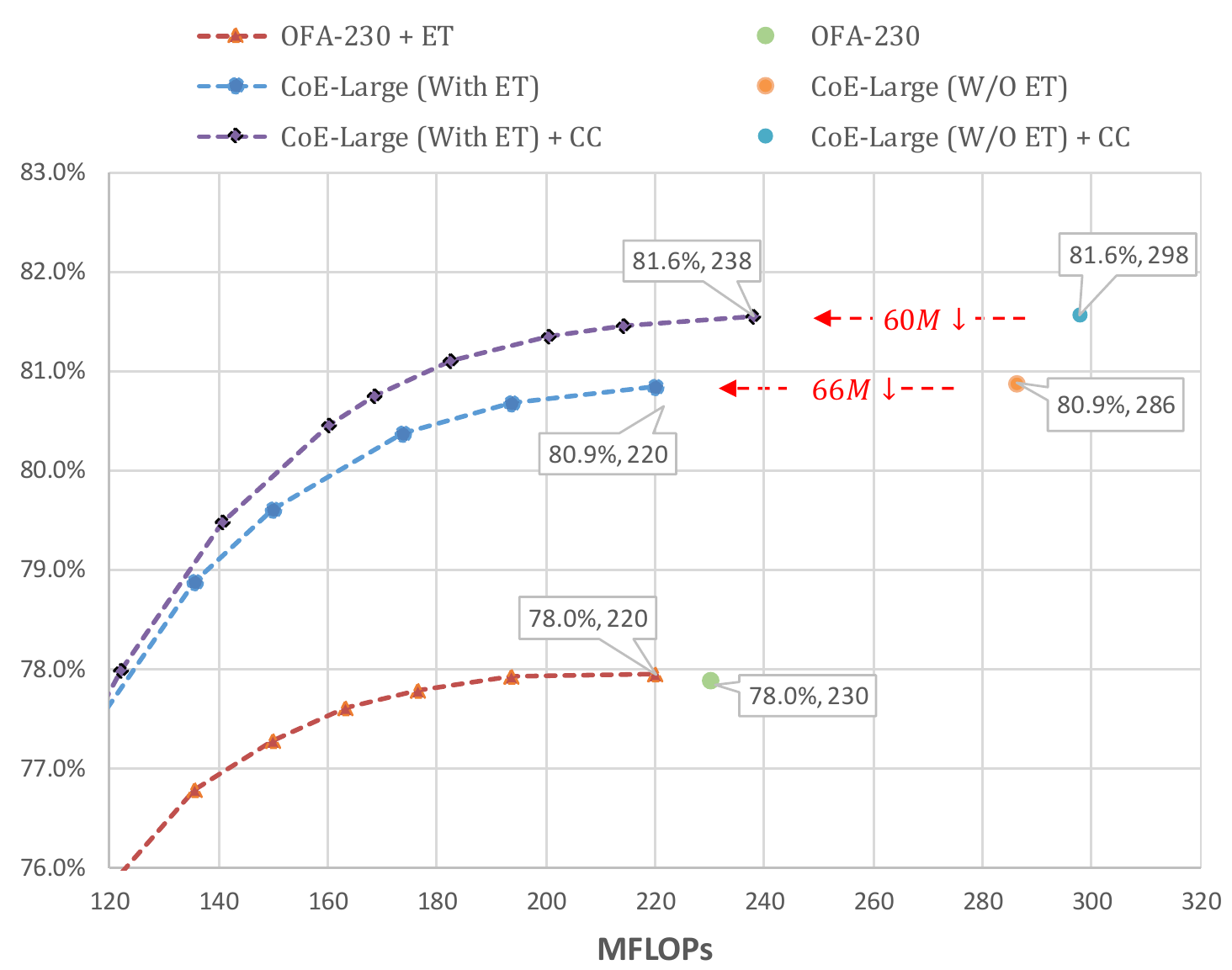}
	\caption{Accuracy v.s. FLOPs. ``ET'' means Early Termination and ``CC'' indicates CondConv.}
	\label{early_termination}
\end{figure}

\subsection{Re-evaluation with Reassessed Labels}
\label{re_evaluation}
As described in paper~\citep{DBLP:journals/corr/abs-2006-07159}, the validation set labels have a set of deficiencies that makes the recent progress on ImageNet classification benchmark suffer from overfitting to the artifacts. To verify the generalization, we use the Reassessed Labels (ReaL)~\citep{DBLP:journals/corr/abs-2006-07159} to re-evaluate our method. The results are shown in Table~\ref{ReaL}. It can be seen that our method still has a remarkable performance, achieving higher accuracy than the compared methods with significantly smaller FLOPs. 

\begin{table}[tbh]
	\footnotesize
	\vspace{-8pt}
	\caption{ReaL and original top-1 accuracy. ``CC'' means CondConv.}
	\label{ReaL}
	\centering
	\begin{tabular}{c|c|c|c}
		\toprule
		\multicolumn{1}{c}{Method}                & FLOPs & ReaL Acc. & Ori. Acc. \\
		\midrule
		OFA-595~\citep{cai2019once}                  & 595M  & 86.0\%    & 80.0\%    \\
		S4L MOAM~\citep{Zhai_2019_ICCV}           & 4B    & 86.6\%    & 80.3\%    \\
		ResNeXt-101~\citep{Xie_2017_CVPR} & 16B   & 85.2\%    & 79.2\%    \\
		ResNet-152~\citep{He_2016_CVPR}           & 11B   & 84.8\%    & 78.2\%    \\
		\midrule
		CoE-Large                                      & 194M  & 86.5\%    & 80.7\%    \\
		CoE-Large + CC                             & 214M  & 86.9\%    & 81.5\%   \\
		\bottomrule
	\end{tabular}
\end{table}

\subsection{Experiment details for the translation task}
\label{trans_details}
Following \citep{attentionis,ott2019fairseq}, CoE-Transformer is trained on the standard WMT 2014 English-German dataset, which has a shared source-target vocabulary of about 37000 tokens. As mentioned, an extra token will be added to this vocabulary. The default training setting is identical with the one described in~\citep{attentionis}, except for the batch size and learning rate becoming larger following~\citep{ott2019fairseq}. Moreover, the parameter $\alpha$ in Eq.\ref{smooth_A} grows linearly from 0.1 to 0.4 with the training going on. We report BLEU on news2014 with a beam width of 4 and length penalty of 0.6 based on a single model obtained by averaging the last 5 checkpoints following~\citep{attentionis,ott2019fairseq}.

\subsection{Analysis of the Learned Expert Selection Patterns}
\label{selection_pattern}

\subsubsection{Expert Selection Patterns when experts share the same architecture}

We have analyzed the selection patterns when experts have different architectures, here we focus on the case that all experts share the same architecture. We adopt the CoE-Large setting with four experts. 

Considering many works~\citep{yan2015hd, mullapudi2018hydranets} select branches based on the category, we firstly experiment to observe the relationship between selection patterns and rough prediction of the delegator on ImageNet validation set. {Based on the predicted class of rough prediction, the validation set can be partitioned into 1000 subsets. Then we calculate the probabilities to select each expert within each subset and get 1000 probability vectors. After clustering, we plot the probability vectors on Fig.\ref{provsclassid}, where each column represents a probability vector.} It can be seen that samples with the same rough prediction class are assigned to different experts. Therefore, we can conclude that the expert is not always selected based on category.

Besides, we further make qualitative analysis on the ImageNet validation set and find some interesting patterns. For example, we find that images predicted as ``meat market'' are most likely to be assigned to the fourth expert if humans are contained. We show those images in Fig.\ref{qualitative_ana}. It can be seen, 27 images are assigned to the fourth expert, among which 22 images contain humans with a ratio of 81.5\%. By contrast, among the 32 images assigned to other experts, only 7 images contain humans with a ratio of 21.9\%. This indicates CoE learns the expert selection patterns automatically, it can be based on properties other than the category.

\begin{figure}[tbh]
	\centering
	\includegraphics[width=0.75\linewidth]{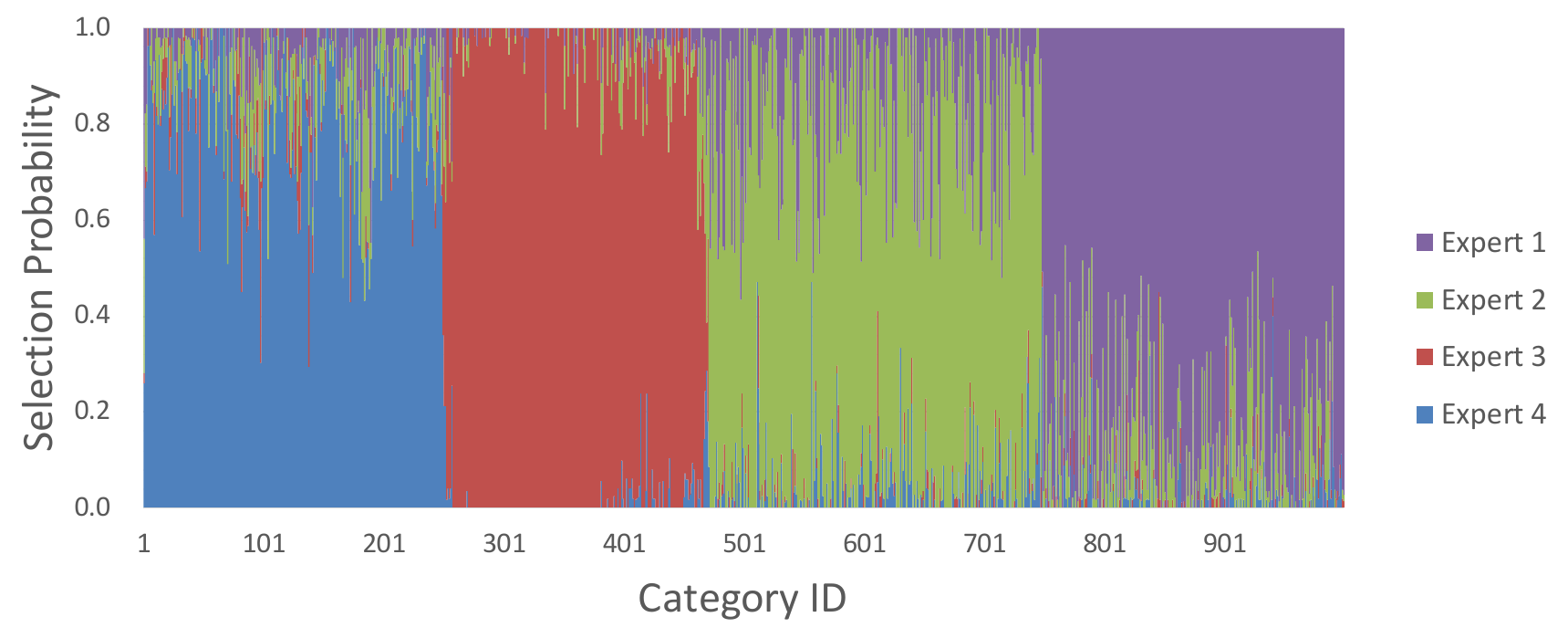}
	\vspace{-10pt}
	\caption{Selection probabilities for each expert. The horizontal axis indicates the rough prediction class. There are a total of 1000 columns, each one represents a probability vector for selecting experts. Probability vectors are clustered for better visualization.}
	\label{provsclassid}
\end{figure}

\begin{figure}[tbh]
	\centering
	\includegraphics[width=1.0\linewidth]{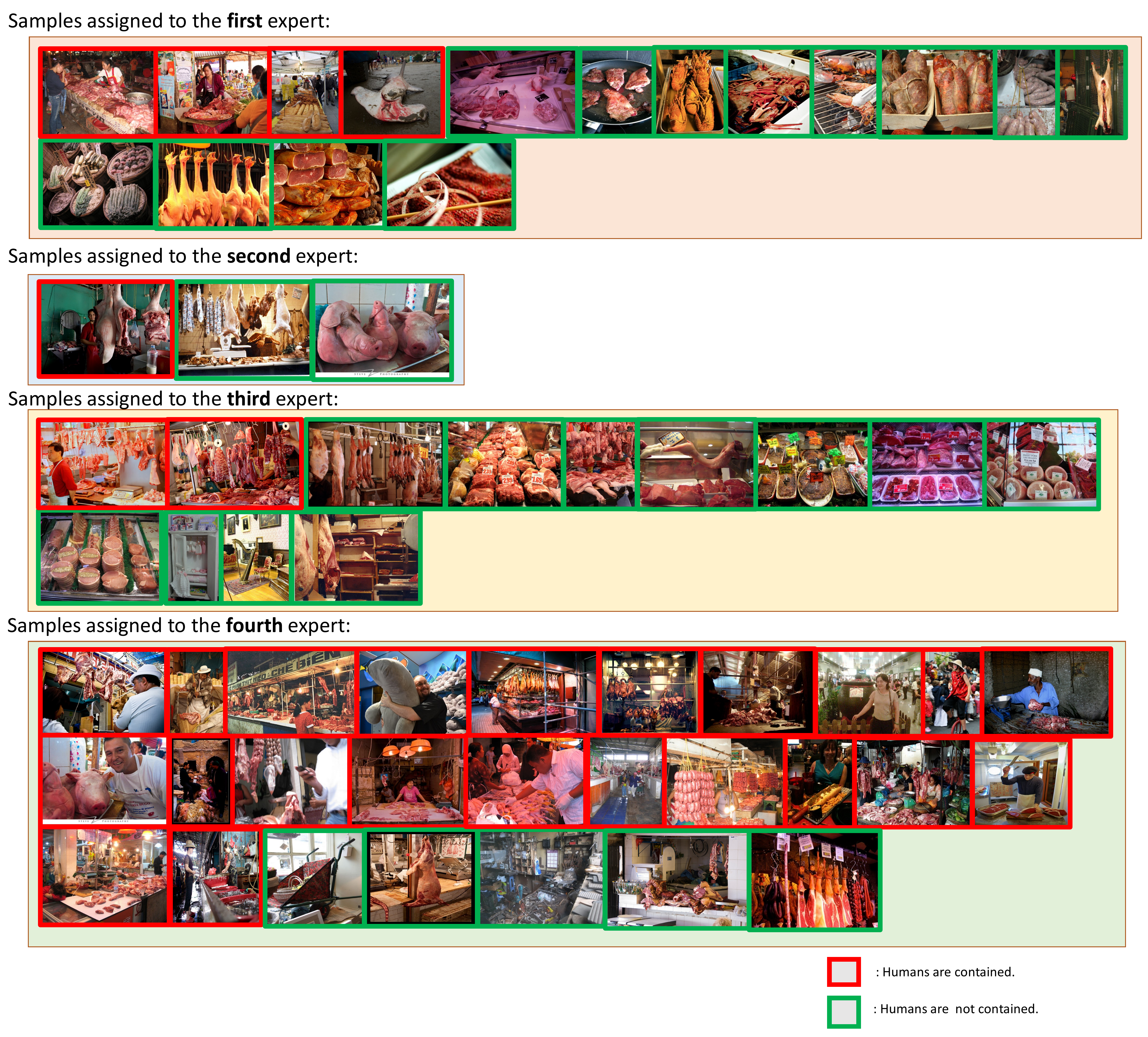}
	\vspace{-10pt}
	\caption{Images that are predicted as 'meat market' by the delegator. They are partitioned into four groups based on which expert is selected. The red border indicates humans are contained, green border indicates humans are not contained.}
	\vspace{-10pt}
	\label{qualitative_ana}
\end{figure}

\end{document}